\documentclass[10pt,twocolumn,letterpaper]{article}

\usepackage{iccv}
\usepackage{times}
\usepackage{epsfig}
\usepackage{graphicx}
\usepackage{amsmath}
\usepackage{amssymb}

\usepackage{bbm}
\usepackage[lined,algonl,boxed]{algorithm2e}

\usepackage[skip=3pt]{subcaption}
\usepackage{multirow}
\usepackage{hhline}
\usepackage{lipsum}
\usepackage{float}
\usepackage{diagbox}
\usepackage{courier}
\usepackage{slashbox,multirow}
\usepackage{enumitem}

\usepackage[utf8]{inputenc}
\usepackage[english]{babel}
\usepackage{amsthm}
\usepackage{microtype}
\newtheorem{remark}{Remark}
\usepackage{framed}
\usepackage{adjustbox}

\makeatletter
\define@key{Gin}{resolution}{\pdfimageresolution=#1\relax}
\makeatother

   {\endMakeFramed}

\AtBeginDocument{%
 \abovedisplayskip=1pt plus 0pt minus 3pt
 \abovedisplayshortskip=1pt plus 0pt minus 3pt
 \belowdisplayskip=3pt plus 0pt minus 2.5pt
 \belowdisplayshortskip=3pt plus 0pt minus 2.5pt

 \floatsep 1.0pt plus 0.0pt minus 3.0pt
 \belowcaptionskip=1pt plus 0pt minus 3pt
 \abovecaptionskip=1.0pt plus 0.0pt minus 3pt
 \dbltextfloatsep=1.0pt plus 0.0pt minus 1.5pt
 \dblfloatsep=1.0pt plus 0.0pt minus 1.5pt
 \intextsep=1.0pt plus 0.0pt minus 2.5pt
 \parskip=1.0pt plus 0.0pt minus 3.5pt
}

\newcommand{\Lap}{\mathbb{L}}
\newcommand{\LapIO}{\overset{\text{\tiny$\leftrightarrow$}}{\mathbb{L}}}
\iccvfinalcopy % *** Uncomment this line for the final submission

\def\iccvPaperID{****} % *** Enter the ICCV Paper ID here

\ificcvfinal\fi

\begin{document}
\setlength{\fboxsep}{0.5pt}
%%%%%%%%% TITLE
\title{Discrete Laplace Operator Estimation for Dynamic 3D Reconstruction}

\author{Xiangyu Xu, 
Enrique Dunn\\
Stevens Institute of Technology, Hoboken, NJ, USA\\
{\tt\small \{xxu24, edunn\}@stevens.edu}
}

\maketitle
\ificcvfinal\thispagestyle{empty}\fi

%%%%%%%%% ABSTRACT
\begin{abstract}
  We present a general paradigm for dynamic 3D reconstruction from multiple independent and uncontrolled image sources having arbitrary temporal sampling density and distribution. Our graph-theoretic formulation models the spatio-temporal relationships among our observations in terms of the joint estimation of their 3D geometry and its discrete Laplace operator. Towards this end, we define a tri-convex optimization framework that leverages the geometric properties and dependencies found among a Euclidean shape-space and the discrete Laplace operator describing its local and global topology.  
  %Our proposal is akin to a dictionary learning instance, where each dictionary atom corresponds to the 3D geometry estimate of each of our 2D observations.
  We present a reconstructability analysis,  experiments on motion capture data and multi-view image datasets, as well as explore applications to geometry-based event segmentation and data association.
\end{abstract}

%%%%%%%%% BODY TEXT
\section{Introduction}
%The task of 
Image-based dynamic reconstruction addresses the modeling and estimation of the spatio-temporal relationships among non-stationary scene elements and the sensors observing them. 
%Research problems within this context include non-rigid structure from motion \cite{tomasi1992shape,bregler2000recovering,xiao2004closed,wu2013towards,dai2014simple,akhter2009nonrigid}, trajectory triangulation \cite{park20103d,park20153d,zhu20113d,zhu2015convolutional,valmadre2012general,zheng2015sparse,zheng2017self,vo2016spatiotemporal,simon2014separable,simon2017kronecker}, image sequencing \cite{basha2012photo,moses2013space,wedge2006motion,padua2010linear,elhayek2012feature,caspi2006feature,gaspar2014synchronization,tuytelaars2004synchronizing,ji2018data}, as well as temporal generalizations of well-established multi-view geometry estimation tasks \cite{vo2016spatiotemporal,albl2017two}. 
This work tackles  estimating the geometry (i.e. the Euclidean coordinates) of a temporally evolving set of 3D points using as input  unsynchronized 2D feature observations with known imaging geometry. 
Our problem, which straddles  both trajectory triangulation and image sequencing,
%is akin to solving the trajectory triangulation problem in the absence of capture sequencing information, and
naturally arises in the context of uncoordinated  distributed capture of an event (e.g. crowd-sourced images or video) and highlights a pair of  open research questions:
%\\ \indent 
 {\em How to characterize and model spatio-temporal relationships among the observations in a data-dependent manner?} %(Neighborhood Volume,  adjacency and affinity)
%\indent
 {\em What role (if any) may  available spatial and temporal priors play within the estimation process?}
The answer to both these questions is tightly coupled to the level of abstraction used to define temporal associations and the scope of the assumptions conferred upon our observations. More specifically, the temporal abstraction level may be quantitative or ordinal (i.e. capture time-stamps vs. sequencing), while the scope of the assumptions may be domain-specific (i.e. temporal sampling periodicity/frequency, choice of shape/trajectory basis) or cross-domain (physics-based priors on motion estimates). 

\begin{figure}[t]
    \centering
    \includegraphics[scale = 0.34, trim={0cm 2.5cm 0cm 1.5cm},clip]{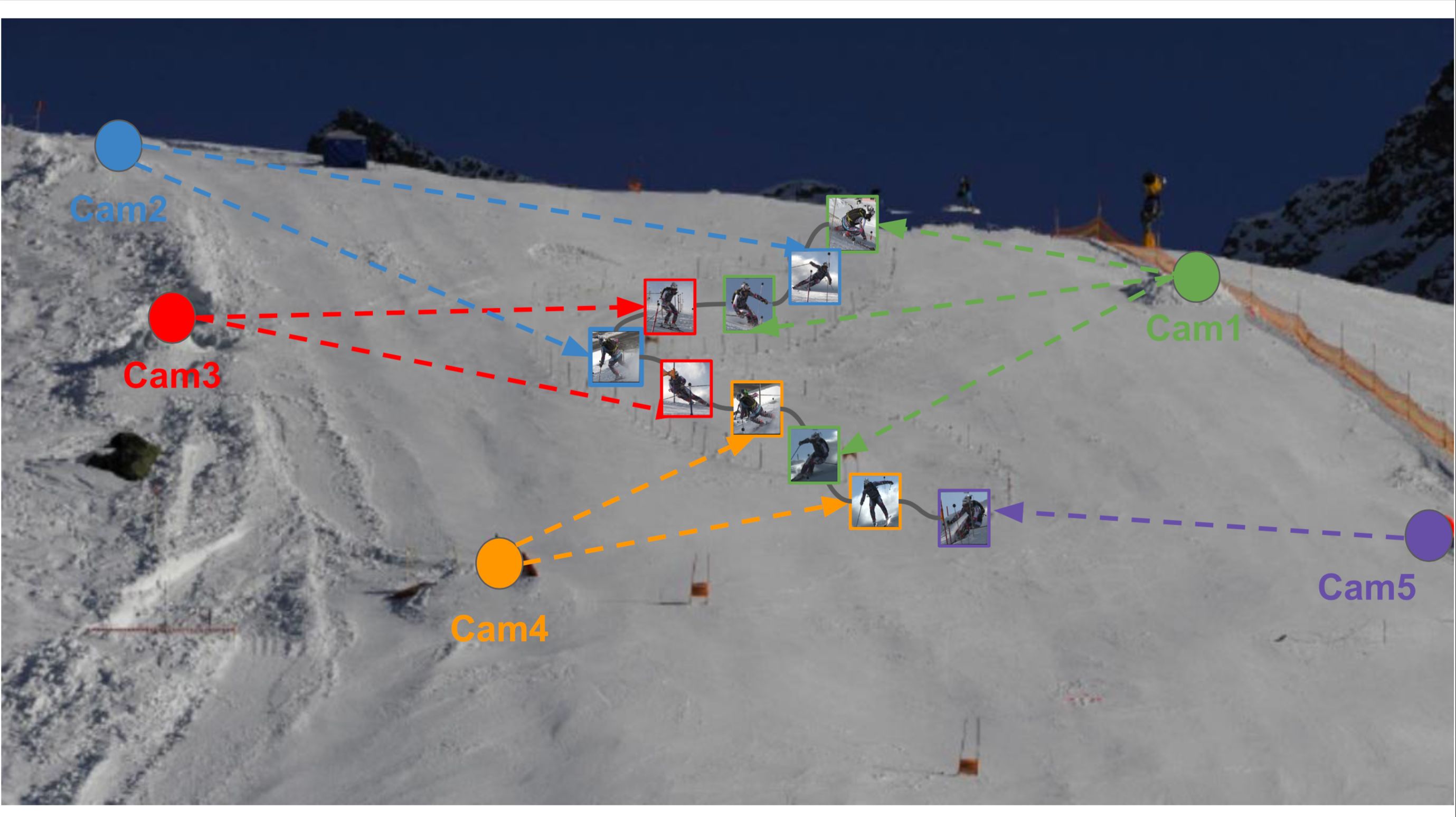}
    \includegraphics[scale = 0.39, trim={0cm 0cm 0cm 0cm},clip]{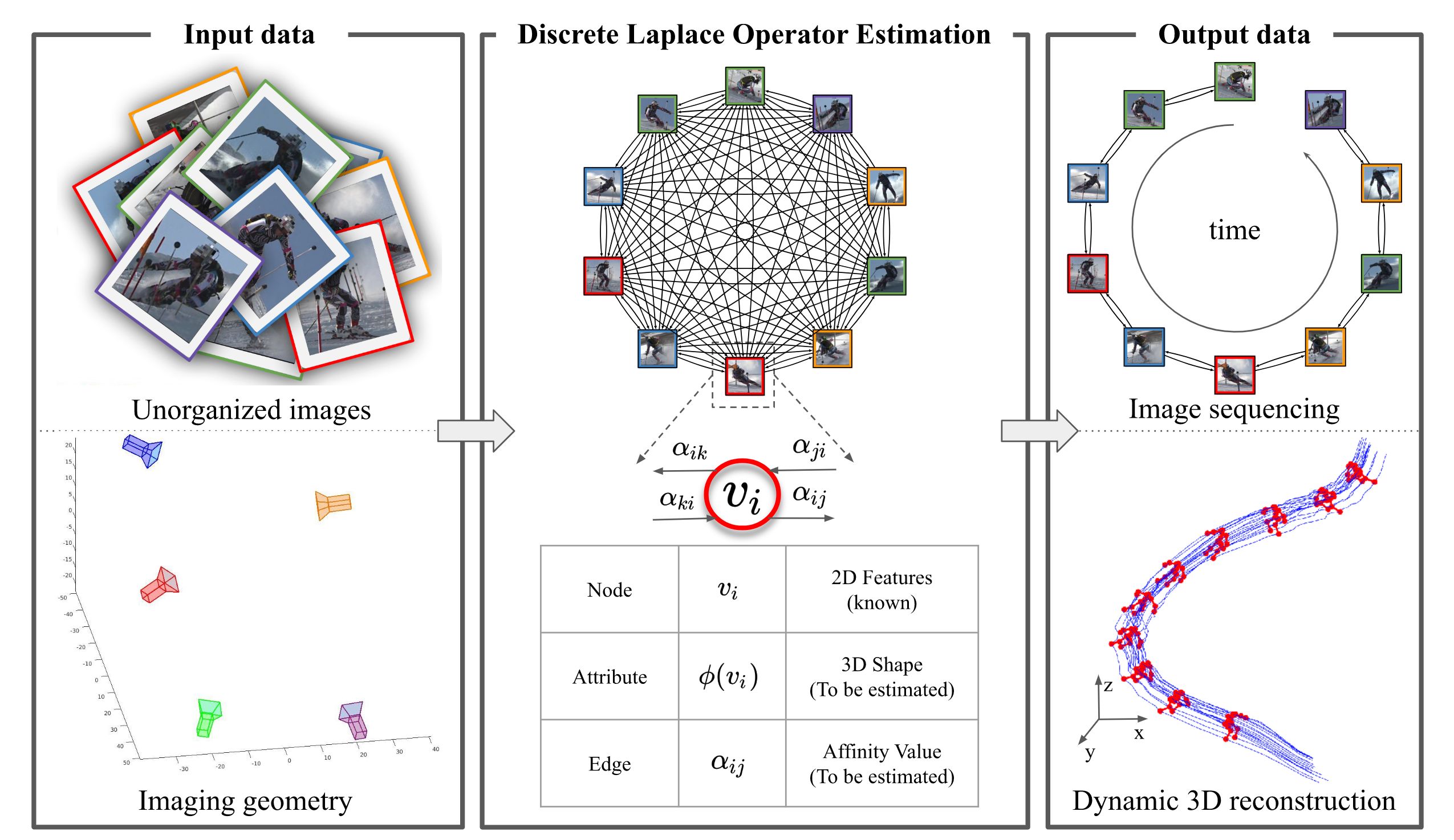}
    \caption{%Framework overview: 
    Multi-view capture produces a set of unorganized 2D observations. Our graph  formulation of dynamic reconstruction jointly estimates  sequencing and 3D geometry.  Imagery adapted from \cite{rhodin2018learning}.}
    \label{fig:Principle}
\end{figure}
%Our framework builds on the following tenets
Estimating either absolute or relative temporal values for our observations would require explicit assumptions on the observed scene dynamics and/or the availability of sampling temporal information  (e.g.  image time-stamps or sampling frequency priors).
In the absence of such information or priors, we strive to estimate
observation sequencing based on  data-dependent adjacency relations defined by a pairwise affinity measure.  Towards this end,
 we make the following assumptions:  {\em \bf A1}) 2D observations are 
%unordered 
samples of the continuous motion of a 3D point set; {\em \bf A2}) the (unknown and arbitrary) temporal sampling density allows 
%a locally linear 3D motion approximation 
approximate local linear interpolation of 3D geometry; 
%for most observations
 and {\em \bf A3})  temporal proximity implies spatial proximity, but not {\em vice-versa} (e.g. repetitive or self-intersecting motion). 
 Under such tenets, we can address multi-view capture scenarios comprised of unsynchronized image streams or the more general case of uncoordinated asynchronous photography.
%Importantly, we use spatial adjacency as a proxy for temporal adjacency, which (as prescribed by our third assumption) is not universally true. 
%Accordingly, we will explore mechanisms for the integration of partial sequencing information priors into our framework. 

We solve a  dictionary learning instance enforcing a discrete differential geometry model, where each dictionary atom corresponds to a 3D estimate, while the set of sparse coefficients describes the spatio-temporal relations among our observations. Our  contributions  are:
\begin{itemize}[noitemsep,topsep=0pt,parsep=0pt,partopsep=0pt]
    \item A graph-theoretic formulation of the dynamic reconstruction problem, where 2D observations are mapped to nodes, 3D geometry are node attributes, and spatio-temporal affinities correspond to graph edges.
    \item The definition and enforcement of spatio-temporal priors, (e.g. anisotropic smoothness, topological compactness/sparsity, and multi-view reconstructability) in terms of the discrete Laplace operator. 
    \item Integration of available per-stream (e.g. intra-video) sequencing info into global ordering priors enforced in terms of the Laplacian spectral signature.
\end{itemize}

%------------------------------------------------------------------------
\section{Related work}
Dynamic reconstruction in the absence of temporal information is an under-constrained problem akin to single view reconstruction \cite{avidan1999trajectory,avidan2000trajectory,han2004reconstruction,shashua1999trajectory,segal20003d,park20113d}. 
Some prior work in trajectory triangulation operate under the assumptions of known sequencing info and/or constrained motion priors.  Along these lines,
Avidan and Shashua \cite{avidan2000trajectory} estimate dynamic geometry  from 2D observations of points constrained to linear and conical motions. 
However, under the assumption of dense temporal motion sampling, the concept of motion smoothness has been successfully exploited \cite{park20103d,park20153d,zhu20113d,zhu2015convolutional,valmadre2012general,zheng2015sparse,zheng2017self,vo2016spatiotemporal,simon2014separable,simon2017kronecker}.
Park et al. \cite{park20103d} triangulate 3D point  trajectories by the linear combination of Direct Cosine Transform trajectory bases with the constraint of a reprojection system. Such a trajectory basis method has low reconstructability when the number of the bases is insufficient and/or  the motion correlation between object and camera is  large. In \cite{park20153d}, Park et al. select number of bases by an N-fold cross validation scheme. Zhu et al. \cite{zhu20113d} apply $L_1$-norm regularization to the basis coefficients to force the sparsity of bases and improve the reconstructability by including a small number of keyframes, which requires user interaction. Valmadre et al. \cite{valmadre2012general} reduce the number of trajectory bases by setting a gain threshold  depending on the basis null-space and propose a  method using a high-pass filter to mitigate low reconstructability for scenarios having no missing 2D observations. Zheng et al. \cite{zheng2017self,zheng2015sparse} propose a dictionary learning method to estimate the 3D shape with partial sequencing info, assuming  3D geometry estimates may be approximated by local barycentric interpolation (i.e. {\em self-expressive} motion prior) and developed a bi-convex framework for  jointly estimating  3D geometry and barycentric weights. However, uniform penalization of  self-expressive residual error and fostering symmetric weight coefficients, handicap the approach against non-uniform density sampling. Vo et al. \cite{vo2016spatiotemporal} present a spatio-temporal bundle adjustment which jointly optimizes camera parameters, 3D static points, 3D dynamic trajectories and temporal alignment between cameras using explicit physics priors, but require frame-accurate initial time offset and low 2D noise. Efforts at developing more detailed spatio-temporal models within the context of NRSFM include \cite{agudo2018deformable, agudo2018scalable, agudo2019robust}.

%, %synchronizing all the video sequencing 
Temporal alignment is a necessary pre-processing step for most dynamic 3D reconstruction methods. Current video synchronization or image sequencing \cite{basha2012photo,moses2013space,wedge2006motion,padua2010linear,elhayek2012feature,caspi2006feature} rely on the image 2D features, foregoing the recovery of the 3D structure. Feature-based sequencing methods like \cite{basha2012photo,wedge2006motion,tresadern2009video} make different assumptions on the underlying imaging geometry. For example, while \cite{basha2012photo} favors an approximately static imaging geometry, \cite{wedge2006motion} prefers viewing configurations with large baselines.
%are limited to static cameras and camera position such as that \cite{basha2012photo} needs cameras to be placed close to each other and \cite{wedge2006motion} prefers to placing cameras far apart.  
Basha et al. \cite{moses2013space} overcomes the limitation of static cameras and improves accuracy by leveraging the temporal info of frames in individual cameras. Padua et al. \cite{padua2010linear} determines spatio-temporal alignment among a partially order set of observation by  framing the problem as  mapping of  $N$ observations into a  single line in $\mathbb{R}^N$, which explicitly imposes a total ordering. %That work illustrated the challenge defining temporal relationship for the case of "slow" 3D motions. 
Unlike previous methods, Gaspar et al \cite{gaspar2014synchronization} propose a synchronization algorithm without tracking corresponding feature between video sequences. Instead, they synchronize two videos by the relative motion between two rigid objects. Tuytelaars et al. \cite{tuytelaars2004synchronizing} 
%try to synchronize two video sequences through their geometry relation in 3D space and recover the 3D scences at the same time. They backproject one tracking point from all frames into space by lines intersection and find the corresponding frames by minimizing line distance. 
determined sequencing based on the approximate 3D intersections of viewing rays under an affine reference frame.
Ji et al.  \cite{ji2016spatio} jointly synchronize a pair of video sequences and reconstruct their commonly observed dense 3D structure by maximizing the spatio-temporal consistency of two-view pixel correspondences across video sequences. 

%------------------------------------------------------------------------
\section{Graph-based Dynamic Reconstruction} \label{sec:framework}
For a set of 2D observations in a single image with known viewing parameters, there is an infinite set of plausible 3D geometry estimates  which are compliant with a pinhole camera  model. We posit that for the asynchronous multi-view dynamic reconstruction of smooth 3D motions, the constraints on each 3D estimate can be expressed in terms of its temporal neighborhood.  That is, we aim to enforce spatial coherence among successive 3D observations without the reliance on instance-specific spatial or temporal models. It is at this point that we come to a {\em chicken-egg} problem, as we need to define a notion of temporal neighborhood in the context of uncontrolled asynchronous capture w/o timestamps or sampling frequency priors. To address this conundrum 
we use spatial proximity as a proxy for temporal proximity, which (as prescribed by our third assumption, i.e. {\bf A3}) is not universally true. Moreover, given that observed events "happen" over a continuous 1D timeline, we would also like to generalize our notion {\em proximity} into one of {\em adjacency}, so as to be able to explicitly define the notion of a local neighborhood.  Towards this end, we pose the dynamic 3D reconstruction problem in terms of discrete differential geometry concepts.
\subsection{Notation and Preliminaries}
\noindent% {\bf Notation } 
%We consider the scenario of $M$ unsynchronized cameras observing $P$ dynamic 3D points, where each camera generates an  image stream $\{\mathcal I\}_m$ where $m = 1,\dots,M$, with an unknown and arbitrary temporal distribution. By considering the aggregation of individual image streams into a single image set $\{\mathcal I\}$, we have $\{\mathcal I\}= \bigcup_{m=1}^{M} \{\mathcal I\}_m$, where the total number of images is given by $N=|\{\mathcal I\}|= \Sigma_{m=1}^{M}|\{\mathcal I\}_m  |$.
We consider 
%the scenario of 
$P$ dynamic 3D points $\{\mathbf{X}_p\}$ observed in $N$ images $\{\mathcal I_n\}$
%Assuming calibrated cameras with known 3D poses,  for each image $\mathcal I_n$  
with known intrinsic and extrinsic camera matrices $\mathbf{K}_n$ and $\mathbf{M}_n$. The 2D observation  of $\mathbf X_p $ in  $\mathcal I_n$ is denoted by $\mathbf{x}_{n,p}$, while its 3D position is denoted by $\mathbf{X}_{n,p}$. 

\noindent {\bf Euclidean Structure Matrix.} The position of all 3D points across all images is denoted by the  matrix 
%Corresponding to each 2D observation, there is a dynamic 3D point $\mathbf{X}_{(p,f)}$. For $N$ cameras with totally $F$ frames and observing $P$ 3D dynamic points, we present our dynamic 3D points matrix as
\begin{equation}
\mathbb{X} = 
\begin{bmatrix}
    \mathbf{X}_{11} & \dots & \mathbf{X}_{1P}  \\
    \vdots & \ddots &\vdots \\
    \mathbf{X}_{N1} & \dots & \mathbf{X}_{NP}  
\end{bmatrix}
\end{equation} 
%where $\mathbf{S}_f$ is a sigle row of the matrix
where each row vector $\mathbf X_{np} \in \mathbb R^3$ specifies  the 3D  Euclidean coordinates of a point.
%We denote by  $\mathbb{X}_{i,:}$ the $i$-th row of the matrix, while $\mathbb{X}_{:,j}$ denotes the $j$-t column. %For an arbitrary vector $\mathbf{x} \in \mathbb{R}^n$, we denote its $i$-th component as  $\left[\mathbf{x}\right]_i$.
Each matrix row $\mathbb{X}_{n,:} \in \mathbb R^{3 \times P}$, represents the 3D shape of the $P$ points in frame $n$. 

%a single point in a Euclidean {\em shape space} (i.e. the aggregated vector of 3D coordinates corresponding to all dynamic points).

%The goal of our dynamic reconstruction  problem is to determine $\mathbb{X}$ given knowledge of a set of sparse 2D observations  $\mathbf x_{p,f}$ and their corresponding perspective projection camera matrices $\mathbf P_f=\mathbf K_f \mathbf M_f$. Moreover, we will explore two scenarios: First, the general case of no available sequencing information for the aggregated image set $\{\mathcal I\}$.\todo{Other experiment?} Second, the case when the sequencing information within each image stream $ \{\mathcal I\}_m$ is known. 
%\footnote{
%Note this does not imply any assumptions on the temporal periodicity or frequency value, and excludes any inter-stream sequencing information.
%}

%-------------m0------------------------------------------------------------
\noindent {\bf Structure Motion Graph.} 
%We consider our temporally evolving 3D geometry as a set of $N$ shapes,  
We define a fully connected graph $G = (V,E)$, 
and map each input image $\mathcal I_n$ to a vertex $v_n \in V$.
%having a one-to-one mapping between our input images $\mathcal I_n \in \{\mathcal I\}$ and our graph vertices $v_n \in V$. 
%We also define a
A multi-value  function   $\phi(\cdot)$ maps a vertex into a point in the {\em shape space}, allowing the interpretation  $\mathbb{X} = \left[\phi(v_1);\dots ; \phi(v_N)\right]$.
%In the absence of any temporal sequencing information, 
Edge weight values $e_{ij} \in E$ are defined by an affinity function $\alpha(\cdot)$ relating points in our shape space, such that $e_{ij}= \alpha_{ij} = \alpha\left(\phi(v_i), \phi(v_j)\right)$.  
%Fig. \ref{fig:Principle} illustrates these relations. %In this work, we model a motion smoothness assumption through the analysis of the discrete differential geometry on the graph. 

%\subsection{Modeling Spatio-Temporal Relationships.}
%-------------------------------------------------------------------------
\noindent {\bf Discrete Laplace operator}. The Laplace operator $\Delta$ is a second  differential operator in  $n-$dimensional Euclidean space, which in Cartesian coordinates  equals to the sum of unmixed second partial derivatives. 
%\todo{Sum of the derivative?Partial derivatives?} 
For a weighted {\em undirected} graph $G = (V,E)$,
the discrete Laplace operator is defined in terms of the Laplacian  matrix:
\begin{equation}
\mathbb{L} = \mathbb{L}_{[\mathbb{A}]} = \mathbb{D} - \mathbb{A} = diag(\mathbb{A} \cdot \mathbf{1}  ) - \mathbb{A}
\label{Laplace}
\end{equation}
where $\mathbb{A}$ is the graph's symmetric affinity matrix, whose values $\mathbb{A}_{ij}$ correspond to the  edge weights $e_{ij} \in \mathbb{R}_{\geq0}$, and $\mathbb{D}$ is the graph's diagonal degree matrix, whose values are the sum of the corresponding row in $\mathbb{A}$.
%See Fig.\ref{fig:Principle} as an example.
\footnote{Alternative definitions  have been used in \cite{fogel2014serialrank,wang2013grassmannian,zeng2018deep,sorkine2005laplacian,zheng2016spectral,chen2007directed,chung2005laplacians}.} 
%The degree matrix $\mathbb{D}$ 
%$= \mathbf {1} \cdot \left[ \mathbf I \bigotimes \left( \mathbb{A} \cdot \mathbf{1} \right) \right] $ 
$\mathbb{L}$ is positive semi-definite, yielding $\mathbf{x}^\top \mathbb{L}\mathbf{x}\geq 0, \;\forall \mathbf{x} \in \mathbb{R}^n$.
When convenient, we obviate the explicit dependence of $\mathbb{L}$ on $\mathbb{A}$. 

\noindent {\bf  Affinity Matrix Decomposition}. The pairwise affinity function $\alpha(\cdot)$ (relating our 3D estimates) is implicitly defined in terms of the estimated entries $\mathbb{A}_{ij}$. Importantly,  these affinity values also encode the graph's local topology (i.e. connectivity). Given the {\em a priori} unknown topology and distribution of our 3D estimates, we make the following design choices: 1) $\mathbb{A}$ is not assumed to be symmetric, yielding a directed  structure graph. 2) we explicitly model the decomposition  $\mathbb{A}=\mathbb{D}\mathbb{W}$, which follows from Eq. (\ref{Laplace}),
\begin{equation}
\mathbb{L} = \mathbb{D} - \mathbb{A} = \mathbb{D}(\mathbb{I}-\mathbb{W})
\label{LapDecomp}
\end{equation}
This decomposition decouples the estimation of each node's degree value  (encoded in $\mathbb{D}$), from the {\em relative} affinity weight values for the node's local neighborhood (encoded in $\mathbb{W}$).

\subsection{Geometric Rationale}

%Accordingly, we will explore mechanisms for the integration of partial sequencing information priors into our framework. 

%$\mathcal{O}\left(\mathbb X, \Theta  \right)$ relating them to our known 2D observations and viewing parameters $\Theta$. 
We leverage the interdependencies among our 3D motion estimates  $\mathbb X$ and its discrete Laplace operator $\mathbb L$, through an optimization framework for their joint estimation.
%It is trivial to construct  $\mathbb L$ given known geometry $\mathbb X$, the functional form of the pairwise affinity $\alpha(\cdot)$, and the graph's connectivity; conversely, $\mathbb X$ can only be attained from $\mathbb L$ through optimization.
  In practice, $\mathbb L$ describes the topology of the given structure  $\mathbb X$ in terms of an affinity function $\alpha(\cdot)$.
The values $\alpha_{ij}$ constitute the  entries of the affinity matrix $\mathbb A_{ij}$ relating the 3D shapes observed at frames $i$ and $j$. These individual values are determined through the estimation of the $\mathbb D$ and $\mathbb W$ variables within our optimization framework. Hence, the affinity $\alpha$ function will not be  explicitly defined, but rather its values will be instantiated from the results of our optimization, which builds upon the following geometric observations.

\begin{remark}[Anisotropic Smoothness  Prior]%Harmonic Motion Prior]
   The norm of the Laplacian's linear form  ($\mathbb {LX}$), tends to vanish when any given function value $\phi(v_i)$  approximates the (affinity-weighted) average of $\phi(v_{j \neq i})$ in its local neighborhood. This follows from the point-wise Laplacian definition
\begin{equation}
\left[\mathbb { LX } \right]_{i,:}=(\Delta\phi)(v_i) = \sum\nolimits_{j}^{N} \mathbb{A}_{ij}\left[\phi(v_i)-\phi(v_j)\right]
\end{equation}
\end{remark}
This implies approximately linear 3D motion segments allow accurate barycentric interpolation from as little as two neighboring 3D motion samples. Conversely, the penalty for poorly approximated non-linear motion segments may be mitigated by the multiplicative contribution of the degree value towards the affinity value, i.e. $\mathbb{A}_{ij} = \mathbb{D}_{ii} \mathbb{W}_{ij}$ 

\begin{remark}[Collapsing Neighborhood Prior]
    The trace of the Laplacian's quadratic form ($\mathbb {X^{\top}LX}$) tends to vanish as  the local neighborhood becomes sparser and more compact, this follows from
\begin{equation}
tr(\mathbb{X}^\top\mathbb{L}\mathbb{X}) = \sum\nolimits_{i,j}^{N}\mathbb{A}_{ij}||\phi(v_i)-\phi(v_j)||_2^2
\label{EqTraceXLX}
\end{equation}
\end{remark}
This implies sparsity in global affinity, while non-zero $\mathbb{A}_{ij}$ values imply proximity among 3D samples $\mathbb{X}_{i,:}$ and $\mathbb{X}_{j,:}$.

\begin{remark}[Spectral Sequencing Prior]
Any line mapping of  $V$ into a vector $\mathbf{f} \in \mathbb{R}^N$   constitutes an ordering of the graph vertices.  Accordingly, when $\mathbf{f}$ is a known and constant affinity preserving mapping, the non-trivial minimization of  $\mathbf{f}^\top\mathbb{L}\mathbf{f}$ will yield  entries in  $\mathbb{L}$  approximating the affinities encoded in $\mathbf{f}$. This follows from 
 \begin{equation}
\mathbf{f}^\top\mathbb{L}\mathbf{f} = \sum\nolimits_{i,j}^{N}\mathbb{A}_{ij}\left(f_i-f_j\right)^2
\end{equation} 
\end{remark}
This implies  enforcing  global sequencing priors by coupling   $\mathbb{L}$'s spectral signature to an input vector $\mathbf f$. 
\begin{figure}[t]
    \centering
    \fcolorbox{black}{black}{\includegraphics[scale = 0.34]{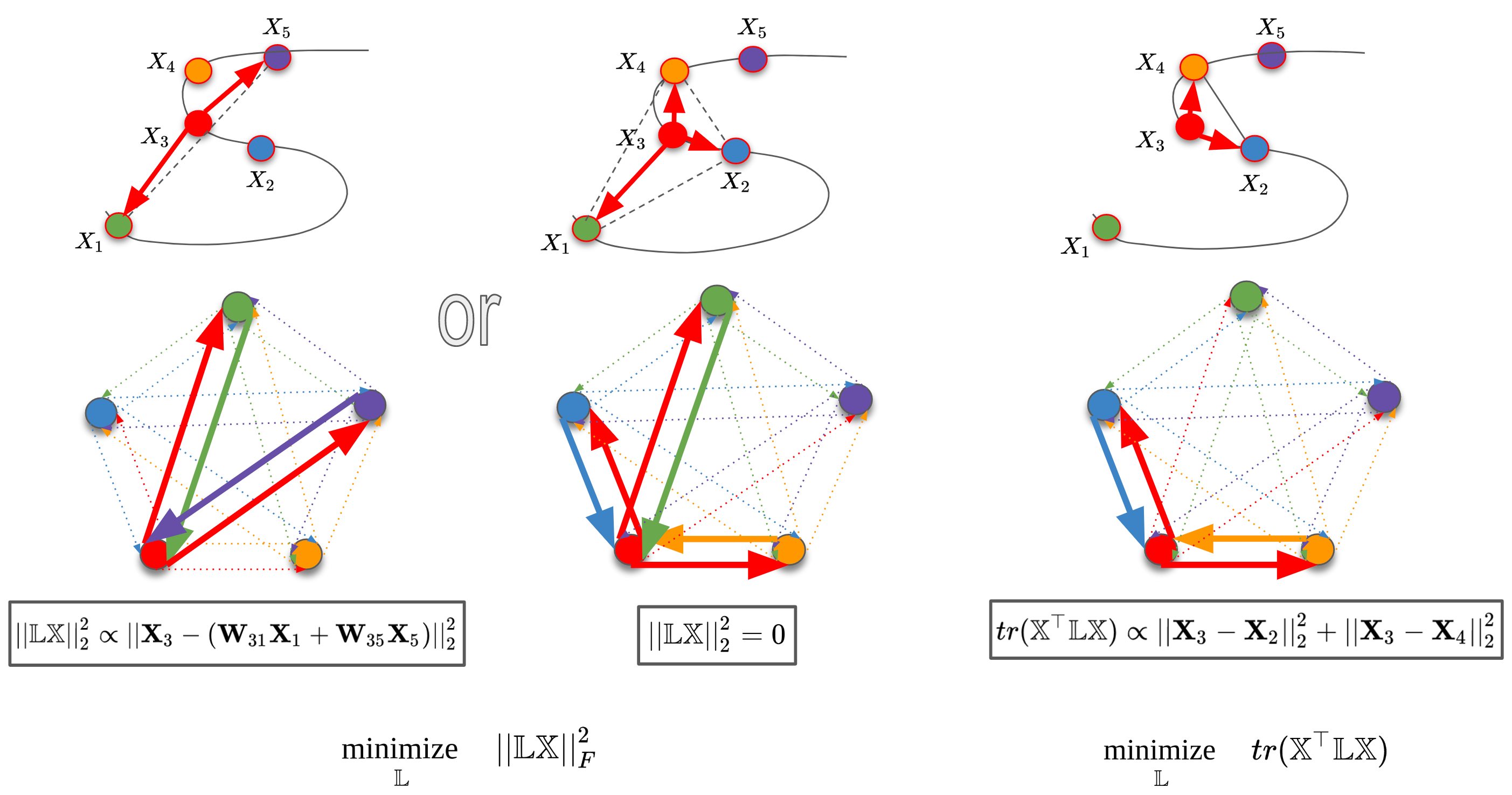}}
    \caption{Geometry of {\bf Remarks 1} \& {\bf 2}. At top: Arrows denote selected neighboring samples and dashed lines their convex hull. At bottom: Corresponding graph edge structure.}
    \label{fig:XLX}
\end{figure}
 \subsection{Optimization Cost Function}
Based on the geometric properties encoded by the discrete Laplace operator the formulate the optimization problem:
\begin{equation}
\underset{\mathbb{X},\mathbb{L}}{\text{min}} \quad \mathcal{S} \left(\mathbb{L} \mathbb{X}\right)+ \mathcal{T} \left(\mathbb{X}^\top\mathbb{L}\mathbb{X}\right) + \mathcal{R}\left(\mathbb L, \Theta  \right) + \mathcal{O}\left(\mathbb X, \Theta  \right),
\label{LapOptOrigXL}
\end{equation}
where 
%$\mathbb{L}$ denotes the motion graph Laplacian, $\mathbb{X}$ the Euclidean structure matrix, and 
$\Theta=\{  \{\mathbf{x}_{np}\},\{\mathbf{K}_n\},\{\mathbf{M}_n\} \}$ denotes the aggregation of all input 2D observations and their camera parameters. Each  cost function term addresses a particular aspect of our optimization. $\mathcal{S}(\cdot)$ fosters local smoothness, $\mathcal{T}(\cdot)$ fosters a linear topological structure, $\mathcal{R}(\cdot)$ fosters strong convergence among viewing rays, while $\mathcal{O}(\cdot)$ reduces reprojection errors. For simplicity, we define the problem variables in terms of $\mathbb{L}$ and  $\mathbb{X}$.
However, given the explicit dependence of $\mathbb{L}$ on $\mathbb{A}$, we'll redefine the joint  optimization of Eq. (\ref{LapOptOrigXL}), as a tri-convex optimization problem over $\mathbb{X}$, $\mathbb{D}$, and  $\mathbb{W}$.

% %------------------------------------------------------------------------

The next two sections  describe  the functional models ($\mathcal{S}$, $\mathcal{T}$,  $\mathcal{R}$, and $\mathcal{O}$) utilized in Eq. (\ref{LapOptOrigXL}), the structure of the estimation variables ($\mathbb{X}$, $\mathbb{D}$, and $\mathbb{W}$), and  the constraints applicable to them. We  present two variants of our general framework,  addressing, respectively, the absence and the estimation of global temporal sequencing priors on the elements of $\{\mathcal I_n\}$.

%We will introduce two methods, one method for reconstructing 3D dynamic structure without knowing the sequencing information using Anisotropic Smoothness Prior and Collapsing Neighborhood Prior mentioned in Sec. \ref{sec:framework} and another method to computing the spectral sequence based 3D structure. Finally we enforce the spectral sequencing prior to improve our 3D structure reconstruction.

%-------------------------------------------------------------------------
\section{Solving for Asynchronous  Photography}
We consider an unordered image set $\{\mathcal I_n\}$, and rely on the {\em Collapsing Neighborhood Prior} to estimate an affinity function matrix whose connectivity approximates a chain-structure connectivity. %Under such graph topology, the 
We interpret such connectivity  as temporal ordering relations among our observations.

%-------------------------------------------------------------------------

\noindent {\bf Enforcing anisotropic smoothness.} 
The functional form
\begin{equation}
\mathcal{S} \left(\mathbb{L} \mathbb{X}\right) = \frac{1}{P}||\mathbb{D}(\mathbb{I}-\mathbb{W})\mathbb{X}||_F^2
\end{equation}
defines the first term of Eq. (\ref{LapOptOrigXL}).
Minimizing $\mathcal{S}$ w.r.t. $\mathbb{X}$ {\em attracts} function values $\phi(v_i)$ towards  the convex hull defined by all $\phi(v_{j \neq i})$ in its local neighborhood. 
Conversely, minimizing $\mathcal{S}$ w.r.t. $\mathbb{L}$ (i.e.  $\mathbb{D}$, $\mathbb{W}$) fosters the {\em selection} of neighboring nodes whose mappings $\phi(v_{j \neq i})$  facilitate barycentric interpolation.
Here, selection refers to assigning non-zero values $\mathbb{A}_{ij}$ in the affinity matrix.

The values in each row of $\mathbb{W}$ (i.e. $\mathbb{W}_{i,:}$) represent the relative affinity weights for $v_i$. Hence, we enforce  1) the sum of each row  equal to 1, and 2) strict non-negativity of all entries in $\mathbb{W}$.Moreover, $\mathbb{D}$ represents the out-degree for each node in the directed graph, akin to a global density estimate. We decouple node degree values from the relative affinity weights in $\mathbb{W}$. We enforce strictly positive degree values $\mathbb{D}_{ii} \geq \epsilon$,  requiring connectivity to at least one adjacent node.

%-------------------------------------------------------------------------

\noindent {\bf Enforcing Neighborhood Locality.} 
For a directed graph, we define the trace of the Laplacian quadratic form   as
\begin{equation}
tr(\mathbb{X}^\top\LapIO\mathbb{X}) = \sum\nolimits_{i,j}^{N}\mathbb{A}_{ij}||\mathbb{X}_{i,:}-\mathbb{X}_{j,:}||_2^2\\
\end{equation}
Where %$\mathbb{L}_{[\mathbb{A}+\mathbb{A}^{\top}]}$ 
$ \LapIO = \Lap_{[\mathbb{A}+\mathbb{A}^{\top}]}$ combines the outdegree and indegree Laplacian matrix, and is compliant with the definition in Eq. (\ref{EqTraceXLX}).
Diagonal entries of the $N \times N$  matrix $\mathbb{X}^\top\mathbb{L}_{[\mathbb{A}+\mathbb{A}^{\top}]}\mathbb{X}$ are the Laplacian quadratic form for each dimension of $\phi()$, 
%$\mathbb X=\phi(V)$, 
and the functional form of $\mathcal T$ in Eq. (\ref{LapOptOrigXL}) is given by their sum:
%The functional form of $\mathcal T$ in Eq. (\ref{LapOptOrigXL}) is 
\begin{equation}\begin{split}
    \mathcal{T} \left(\mathbb{X}^\top\mathbb{L}\mathbb{X}\right) = \frac{\lambda_1}{P}\sum\nolimits_{i,j}^{N}\mathbb{D}_{ii}\mathbb{W}_{ij}||\mathbb{X}_{i,:}-\mathbb{X}_{j,:}||_2^2
    \label{EqT_XLX}
\end{split}\end{equation}

Minimizing  $\mathcal{T}$ w.r.t. $\mathbb{X}$ (i.e. fixing  $\mathbb{A}$) {\em attracts} the estimates $\phi(v_{j \neq i})$ of neighboring elements to be {\em near} to $\phi(v_{i})$.
Conversely, minimizing $\mathcal{T}$ w.r.t. $\mathbb{A}$, fosters the {\em selection} of nearby nodes to form a compact neighborhood, as defined by the weighted sum of the magnitude of the difference vectors  $\phi(v_i) - \phi(v_{j \neq i})$, $ \forall \mathbb{A}_{ij}\neq 0$.   

%-------------------------------------------------------------------------
%\subsubsection{Viewing ray constrains}
\noindent {\bf Enforcing Observation Ray Constrains.} 
We penalize the distance of a 3D point $\mathbb X_{np}$ to its known viewing ray using 
%\begin{equation}
  $ \mathbf{d}_{np} = ||(\mathbf{X}_{np} - \mathbf{C}_{n})\times\mathbf{r}_{np}||_2,$
%\end{equation}
where $\mathbf{r}_{np}$ is a unit vector parallel to the viewing ray  $ \mathbf{R}_n^\top \mathbf{K}_n^{-1}[\mathbf{x}_{np}^\top \quad 1]^\top $   and  camera pose parameters are given by $\mathbf{M}_n = [\mathbf{R}_n|-\mathbf{R}_n\mathbf{C}_n]$ \cite{zheng2017self}. 
%and $\mathbf{K}_n$ are the intrinsic parameters. 
The functional form of $\mathcal{O}$ from Eq. (\ref{LapOptOrigXL}) is 
\begin{equation}
    \mathcal{O}\left(\mathbb X, \Theta  \right) = \sum\nolimits_{n,p}^{N,P} \frac{\lambda_2}{NP}||\mathbf{d}_{np}||_2^2 ,
\end{equation}
which is  quadratic  for $\mathbb{X}$. The value of $\lambda_2$ depends on the 2D noise level  and the mean camera-to-scene distance.\\
%, since it control how much the 3D point could deviate from the viewing ray.
\noindent {\bf Enforcing Multi-view Reconstructability}. 
Viewing geometry plays a determinant role in the overall accuracy of our 3D estimates (see section \ref{SecRecAcc} for a  detailed analysis). Intuitively, for moderate-to-high 2D noise levels, the selection of temporally adjacent cameras with small baselines will amplify 3D estimation error. In order  to foster the selection of cameras having favorable convergence angles among  viewing rays corresponding to the same feature track, we define the functional form of $\mathbb R$ from Eq. (\ref{LapOptOrigXL})  as
\begin{equation}
    \mathcal{R}\left(\mathbb L, \Theta  \right) = \frac{\lambda_3}{NP} \sum\nolimits_{i,j,p}^{N,N,P} \left(\mathbb D_{ii} \mathbb W_{ij} \left(\mathbf r_{ip} \cdot \mathbf r_{jp} \right)\right)^2
\end{equation}
%-------------------------------------------------------------------------
\section{Solving for Unsynchronized Image Streams} \label{sequencing}
\begin{figure}[t!]
    \centering
	\fbox{\begin{subfigure}[t!]{0.23\textwidth}
	    \centering
		\includegraphics[scale = 0.19]{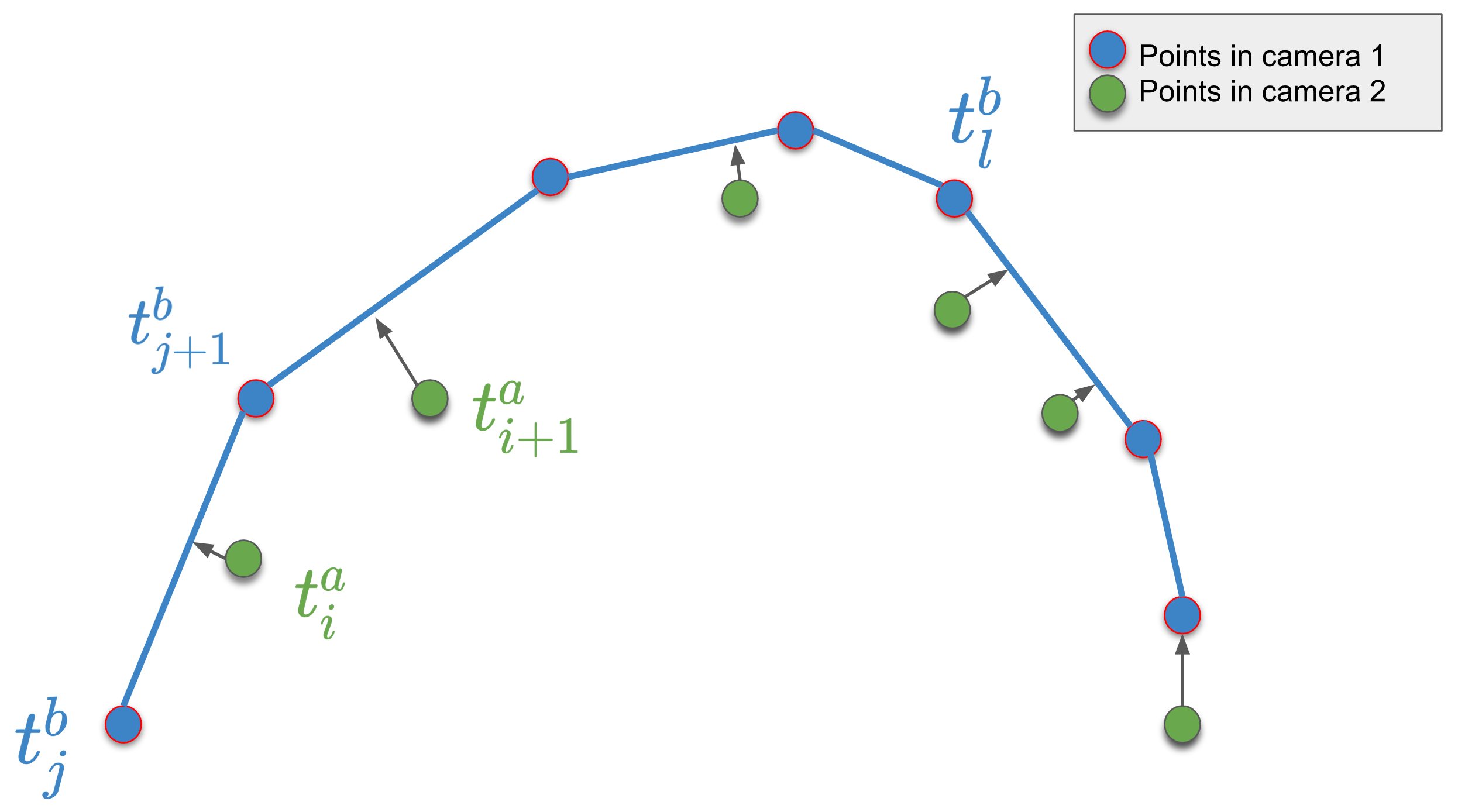}
		\caption{Matching with DTW.} 
		\label{fig:DTW}
	\end{subfigure}
	%\vspace{2em}
	\begin{subfigure}[t!]{0.23\textwidth}
	    \centering
		\includegraphics[scale = 0.19]{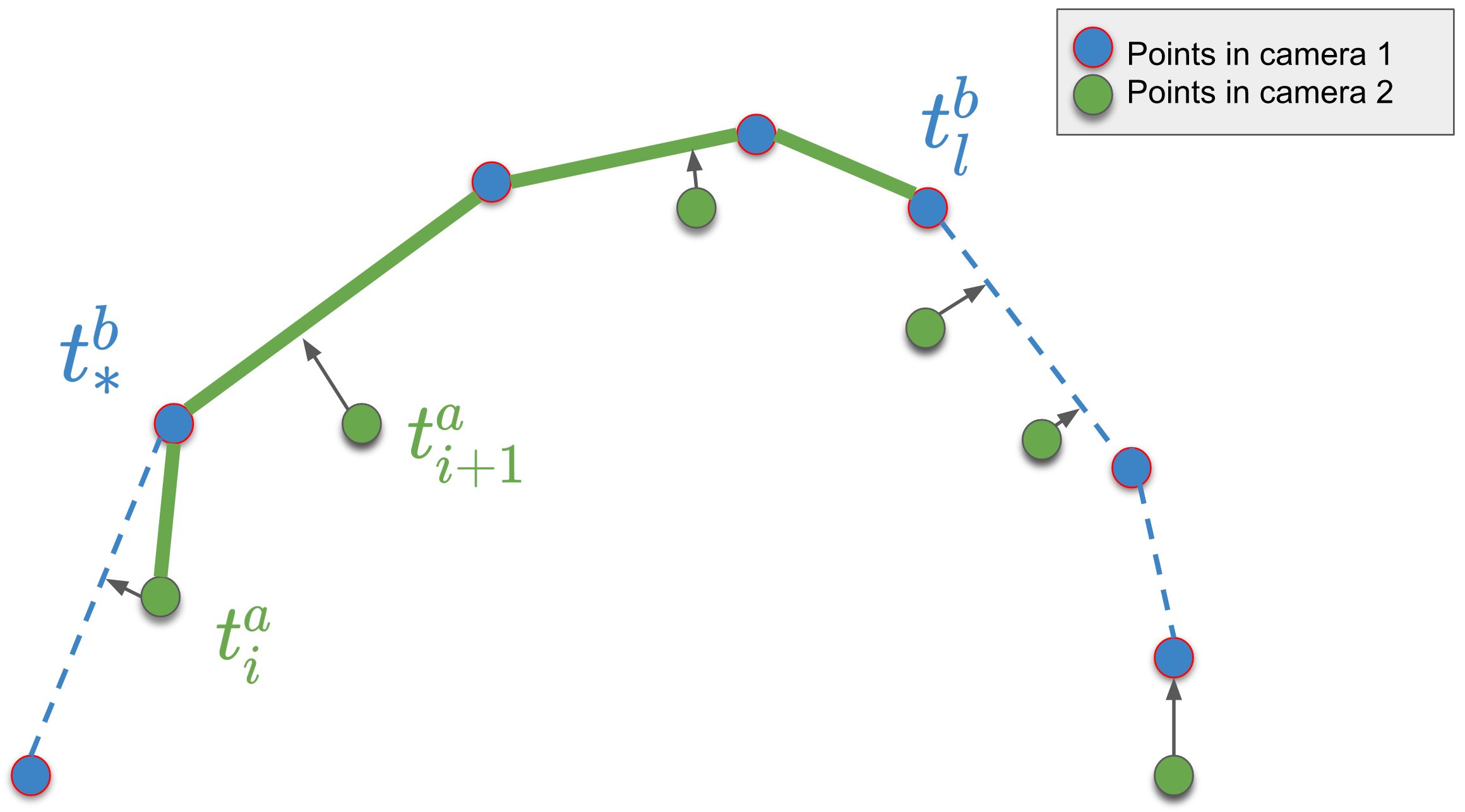}
		\caption{Arc distance} 
		\label{fig:ArcDistDiff}
	\end{subfigure}}
	\caption{Arc distance between two observations of the same 3D point across different image streams.} 
\end{figure}

Given an image set comprised of the aggregation of multiple image streams, we ascertain partial sequencing (i.e. within disjoint image subsets).  We use this  info in two different ways: First, we enforce spatial smoothness among successive observations from a common stream. Second, {\em we integrate  disjoint local sequences into a global sequencing estimate we enforce through our optimization}. \\
%We have advocated the joint estimation of structure and its affinity function. In this section 
\noindent {\bf Enforcing Intra-Sequence Coherence}.
We define $\mathbb{W} = \mathbb{W}_{var} + \mathbb{W}_{prior}$, where $\mathbb{W}_{var}$ constitutes the variable component of our estimation, while $\mathbb{W}_{prior}$  encodes small additive values for the immediately prior and next frames from the same image stream. The collapsing neighborhood prior will enforce such {\em pseudo-adjacent} 3D estimates to be similar. \\
\noindent {\bf Manipulating the Spectral Signature of $\mathbb{L}$}.
For a given global sequencing prior, in the form of a line embedding $\mathbf f \in \mathbb R^N$ of all our graph nodes, we modify  Eq. (\ref{EqT_XLX}) to be
\begin{equation}\begin{split}
%\textstyle 
%\mathcal{T}\left(\mathbb{X}^\top\mathbb{L}\mathbb{X}\right) \Rightarrow
    \mathcal{T} \left(\mathbf{f}^\top\mathbb{L}\mathbf{f}\right) = \frac{\lambda_1}{P}\sum\nolimits_{i,j}^{N}\mathbb{D}_{ii}\mathbb{W}_{ij}\left(\mathbf{f}_{i}-\mathbf{f}_{j}\right)^2.
\end{split}\label{T-fLf}\end{equation}
We now describe how we determine such line embedding $\mathbf f$. \\
\noindent {\bf Integrating Global Sequencing Priors}.
Our goal is to integrate preliminary (e.g. initialization) geometry estimates, $\mathbb{X}^{init}$, with reliable but partial sequencing information (e.g. single video frame sequencing) into a global sequencing prior. Towards this end, we pose image sequencing from a given 3D structure $\mathbb{X}$ as a dimensionality reduction instance,  where the goal is to find a line mapping which preserves (as much as possible) pairwise proximity relations among 3D estimates. 
%In this context, we focus on defining a pairwise relation function. 
While using  Euclidean distance as a pairwise proximity measure is suitable for approximately linear motion,  non-linear motion manifolds (i.e. repetitive or self-intersecting motions) may collapse temporally distant observations to proximal locations in the line embedding. \\
\noindent {\bf Arc Distance through Dynamic Time Warping}. We define approximate 3D trajectory {\em arc distance} for shapes within sequenced images streams, as the sum of 3D line segment lengths among adjacent observations, see Fig. \ref{fig:DTW}. To generalize this notion across  image streams, we perform global approximate inter-sequence registration through Dynamic Time Warping (DTW). Our goal is to assign to each 3D estimate $t^a_i$ along trajectory $a$ the closest line segment $(t^b_j,t^b_{j+1})$  in each of the other trajectories $b \neq a$, without violating any sequencing constraints in our assignments, which we define 
\begin{equation}
    t^a_i \rightarrow (t^b_j,t^b_{j+1}) \quad \nexists \quad t^a_{k > i} \rightarrow (t^b_{l<j}, t^b_{l+1}) \;\; \forall a \neq b
\end{equation}
Once all assignments are made, inter-sequence arc-length between $t^a_i$ and $t^b_l$ is trivially computed as the sum of 1) distance to the element $t^b_*$ in the line segment $(t^b_j, t^b_{j+1})$ closest to the $t^b_l$, plus 2) the intra-sequence arc distance between $t^b_{*}$ and $t^b_{l}$.  Fig. \ref{fig:ArcDistDiff} illustrates the arc distance  from points between $t^a_i$ and $t^b_l$ as the length of green line.

%\subsubsection{Dimensionality reduction}
\noindent {\bf Dimensionality Reduction Methods.} 
We use arc length to define a pair-wise distance matrix ${\mathbb Z}$, from which we 
attain a  vector embedding $\mathbf f \in \mathbb R^n$ through Spectral Ranking (SR) \cite{fogel2014serialrank,chung1997spectral} and Multidimensional Scaling (MDS) \cite{abdi2007metric}.
%Spectral Ranking \cite{fogel2014serialrank} \cite{chung1997spectral} which computes $\mathbf{f}$ by setting the frames with small distance as close as possible and Multidimensional Scaling \cite{abdi2007metric} which computes $\mathbf{f}$ by preserving the pairwise distance. 
Sequencing is attained by sorting $\mathbf f$. 
Alternatively, we  interpret ${\mathbb Z}$ as a complete graph's weight matrix and find the approximate shortest Hamiltonian path (SHP). %However, it's a NP-complete problem which cannot be solved in polynomial time. 
Table \ref{tab:Kendall} compares  these  methods operating on ${\mathbb Z}$ and the Euclidean distance matrix ${\mathbb{Z}^E}$, both  matrices were computed from  $\mathbb{X}^{init}$ and $\mathbb{X}^{opt}$, which denote respectively,  the  initial 3D structure
%obtained through pseudo-triangulation 
and the estimated 3D structure after our optimization.

\begin{table}[t!]
\centering
\resizebox{\columnwidth}{!}{%
    \begin{tabular}{|c|c|c|c|c|c|c|c|}
         \hline
         \multicolumn{2}{|c|}{} & \multicolumn{2}{c|}{Linear motion} & \multicolumn{2}{c|}{Nonlinear motion} & \multicolumn{2}{c|}{Repeating motion}\\ 
         \hhline{~~------}
         \multicolumn{2}{|c|}{} & $\mathbb{X}^{init}$ & $\mathbb{X}^{opt}$ & $\mathbb{X}^{init}$ & $\mathbb{X}^{opt}$ & $\mathbb{X}^{init}$ & $\mathbb{X}^{opt}$ \\
         \hline
         \multirow{2}{*}{SR} & $\mathbb{Z}^E$ & 0.9956 & 0.9996 & 0.9807 & 0.9991 & 0.6754 & 0.7140\\
         \hhline{~-------}
          &$\mathbb{Z}$& 0.9965 & 1 & 0.9570 & 1 & 0.9711 & 0.9934\\
         \hline
         \multirow{2}{*}{MDS}&$\mathbb{Z}^E$& 0.9943 & 1 & 0.7614 & 0.7044 & 0.6421 & 0.6553\\
         \hhline{~-------}
         &$\mathbb{Z}$& 0.9961 & 1 & 0.8741 & 1 & 0.9316 & 0.9732\\
         \hline
          \multirow{2}{*}{SHP}&$\mathbb{Z}^E$& 1 & 1 & 0.4368 & 0.9996 & 0.3329 & 0.7912\\
         \hhline{~-------}
         &$\mathbb{Z}$& 1 & 1 & 0.5325 & 0.9996 & 0.3947 & 0.7934\\
         \hline
    \end{tabular}
    }
    \caption{Kendall rank correlation {\em vs.} ground truth ordering for sequencing attained from initial and estimated structure.}
    \label{tab:Kendall}
\end{table}

%------------------------------------------------------------------------
\section{Optimization} \label{Sec:Optimization}
Eq. (\ref{LapOptOrigXL}) is a tri-convex function for variable blocks $\mathbb{X}$, $\mathbb{W}$ and $\mathbb{D}$. 
%There some biconvex optimization methods for different condition, such as Alternate Convex Search(ACS), Jointly Constrained Biconvex Programming \cite{gorski2007biconvex} and Global Optimization Algorithm(GOA) \cite{floudas1990global}. 
We use the ACS \cite{gorski2007biconvex} strategy,  alternatively optimizing over each variable block while fixing the other two. For the first iteration, we initialize $\mathbb{D}$ and $\mathbb{X}$ (to be described), then we alternatively optimize over each variable blocks in the order of $\mathbb{W}$, $\mathbb{D}$ and $\mathbb{X}$ until (thresholded) convergence  of  our cost function among successive iterations.

%-------------------------------------------------------------------------
\noindent {\bf Optimizing over $\mathbb{X}$}. 
While variable blocks $\mathbb{W}$ and $\mathbb{D}$ are fixed, the cost function (\ref{LapOptOrigXL}) is a quadratic equation for block $\mathbb{X}$ without any constraints. The solution for this quadratic programming problem is the set of variable values found at the zeros of the derivative of the cost function.

%-------------------------------------------------------------------------
\noindent {\bf Optimizing over $\mathbb{W}$.} 
With $\mathbb{X}$ and $\mathbb{D}$ fixed, minimizing $\mathcal{S}  \left(\mathbb{L} \mathbb{X}\right)$, $\mathcal{T} \left(\mathbb{X}^\top\mathbb{L}\mathbb{X}\right)$, $\mathcal{O}\left(\mathbb X, \Theta  \right)$ and $\mathcal{R}\left(\mathbb L, \Theta  \right)$, respectively, yield a quadratic equation, linear equation and constant value for $\mathbb{W}$,  making the cost function a quadratic equation for $\mathbb{W}$ 

\begin{equation}
\begin{aligned}
 & \underset{\mathbb{W}}{ \text{min}}%\;\; 
 & & \frac{1}{P}|| \mathbb{D}( \mathbb{I}- \mathbb{W})\mathbb{X}||_F^2 +  \frac{\lambda_1}{P}\sum\limits_{ij}^N\mathbb{D}_{ii}\mathbb{W}_{ij}||\mathbb{X}_{i,:}-\mathbb{X}_{j,:}||_2^2  \\
 &&& + \frac{\lambda_3}{NP} \sum\nolimits_{i,j,p}^{N,N,P} \left(\mathbb D_{ii} \mathbb W_{ij} \left(\mathbf r_{ip} \cdot \mathbf r_{jp} \right)\right)^2 \\
 & \text{s.t.} %\;\; &  
 & & \mathbb{W} \mathbf{1}_{N \times 1} =  \mathbf{1}_{N \times 1}, \quad  \mathbb{W} \geq 0
\label{CostFunForW}
\end{aligned}
\end{equation}

Each row of $\mathbb{W}$ is independent and is solved as a quadratic programming problem with linear constrains. We optimize each row in parallel by the  Active-Set method  in \cite{chen2014fast}. 

%-------------------------------------------------------------------------
\noindent {\bf Optimizing over $\mathbb{D}$.}
%Optimizing over $\mathbb{D}$ is similar to $\mathbb{W}$. 
When $\mathbb{X}$ and $\mathbb{W}$ are fixed, optimizing  Eq. (\ref{LapOptOrigXL}) yields a quadratic equation in terms of the diagonal values of $\mathbb{D}$. We optimize the same equation as Eq. (\ref{CostFunForW}), but with  linear constrains $\{tr(\mathbb{D}) = 1, \mathbb{D} \geq 0\}$,  normalizing the outdegree sum to one.

%-------------------------------------------------------------------------
\noindent {\bf Optimizing for the spectral sequencing prior} \label{Sec:fLf}
When optimizing over $\mathbb{W}$ or $\mathbb{D}$, 
%the second term in Eq. (\ref{CostFunForW}) is equivalent to computing $\mathbb{L}$ based on the pairwise Euclidean distance of elements in our shape space. However, 
the matrix $\mathbb{X}$ is  replaced by a vector  $\mathbf{f}$,  computed from the current estimate of  $\mathbb{X}$, through one of the dimensionality reduction methods described earlier (e.g. MDS applied to $\mathbb Z$)  %external method  Sec.\ref{sequencing} which will have better adjacent information. 
Hence, the second term becomes
\begin{equation}
\mathbf{f}^\top\LapIO\mathbf{f} = \sum\nolimits_{ij}^N\mathbb{D}_{ii}\mathbb{W}_{ij}(\mathbf{f}_{i}-\mathbf{f}_{j})^2\\
\label{fLf}
\end{equation}
When using MDS as the dimensionality reduction method, $\mathbf{f}$ approximately preserves the pairwise Arc distance, allowing direct implementation within  Eq. (\ref{fLf}). When using SR, $\mathbf{f}$ corresponds to the graph's Fiedler vector, whose entry values range from -1 to 1; requiring a uniform scaling in order to match the range of the current structure estimate $\mathbb{X}$.
%, so we choose MDS in the experiment.

%-------------------------------------------------------------------------
\noindent {\bf Initialization.} \label{Initial}
We initialize  the degree matrix to be $\mathbb D_{ii}=1/N$. 
%To initialize  $\mathbb X$, we assume the absence of concurrent observations. Accordingly, 
We  initialize the 3D structure $\mathbb{X}^{init}_{n,:}$ observed in $\mathcal I_{n}$ by the approximate two-view pseudo-triangulation of each viewing ray $\mathbf{r}_{np}$ with its corresponding viewing ray $\mathbf{r^*}_{m \neq n,p}$ from the {\em most convergent}  image $\mathcal I_{m}$, which is the $\mathcal I_{m}$ with the minimum aggregated pseudo-triangulation error when considering all commonly observed points. 

\section{Structure Reconstruction Accuracy. \label{SecRecAcc}}
\begin{figure*}[]
    \centering
    \fbox{\begin{subfigure}[ht!]{0.2\textwidth}
	    \includegraphics[scale = 0.28,trim={0cm 0cm 0cm 0cm},clip]{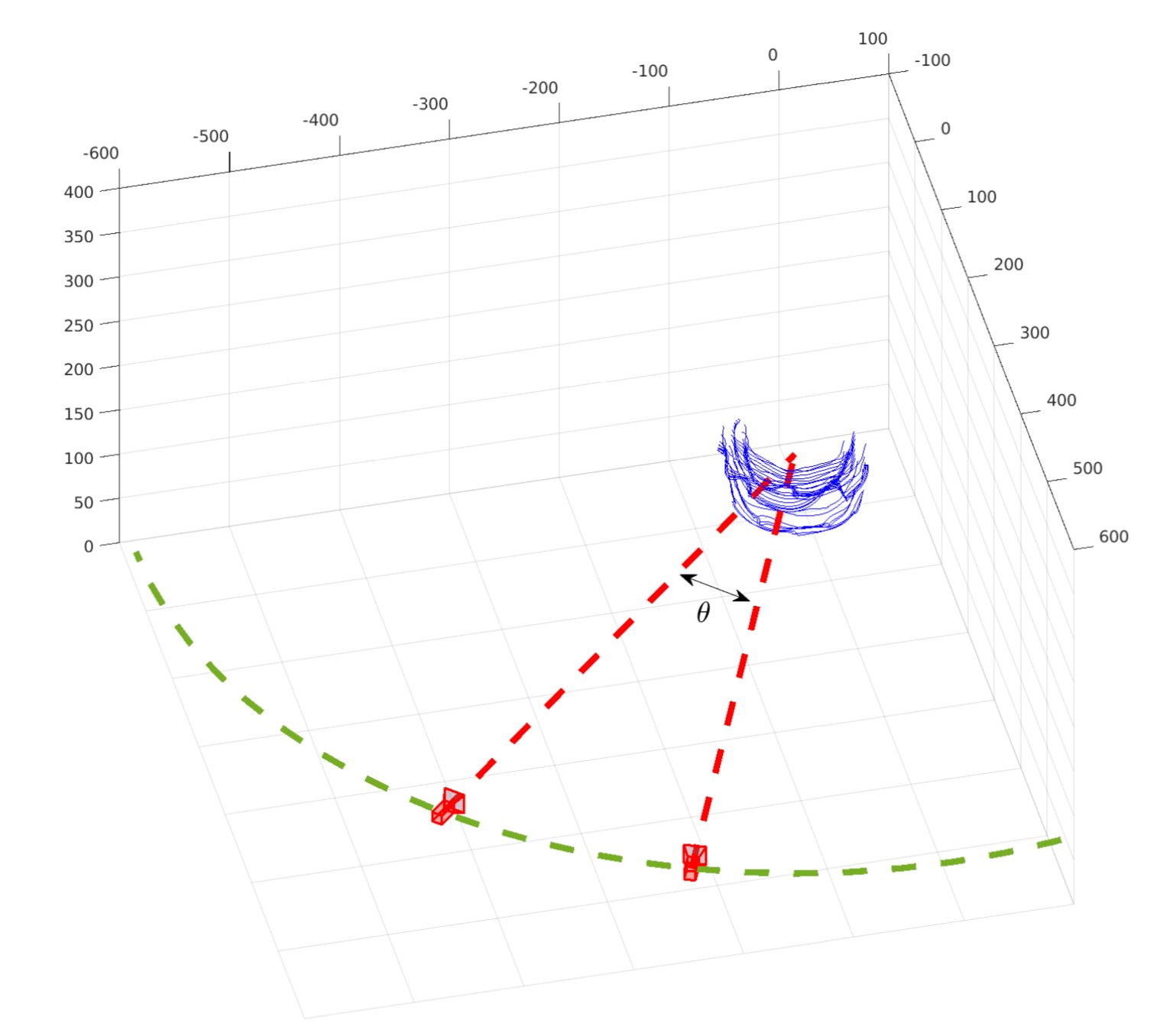}
	    \caption{Convergence angle $\theta$}
     	\label{fig:Angle between cameras}
	\end{subfigure}
	\hspace{1.5em}
	\centering
	\begin{subfigure}[ht!]{0.22\textwidth}
	    \includegraphics[scale = 0.22]{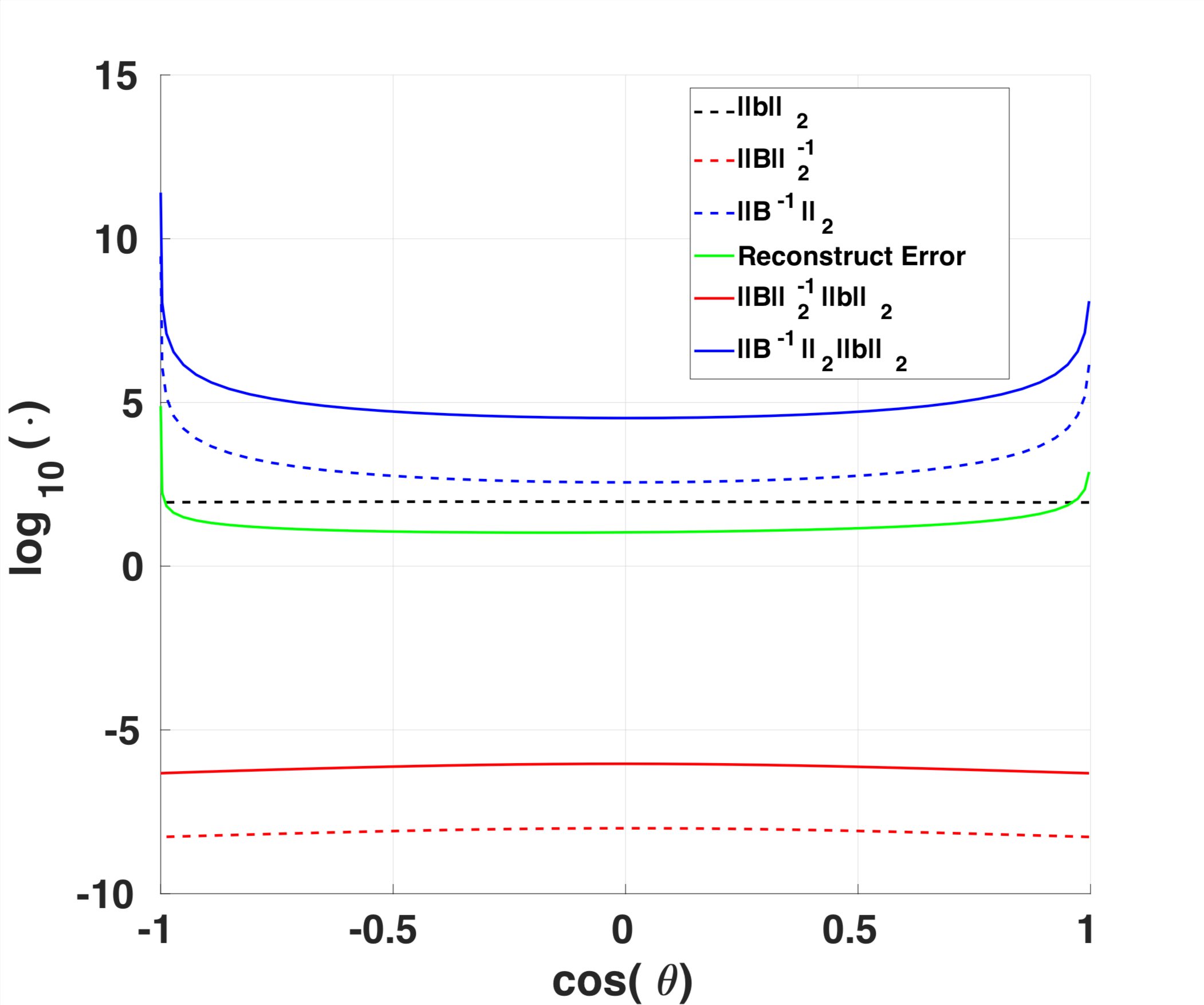}
	    \caption{$\mathbb B$ term analysis}
     	\label{fig:Bterm}	    
    \end{subfigure}}
    \hspace{2em}
    \centering
    \fbox{\begin{subfigure}[ht!]{0.2\textwidth}
	    \includegraphics[scale = 0.23]{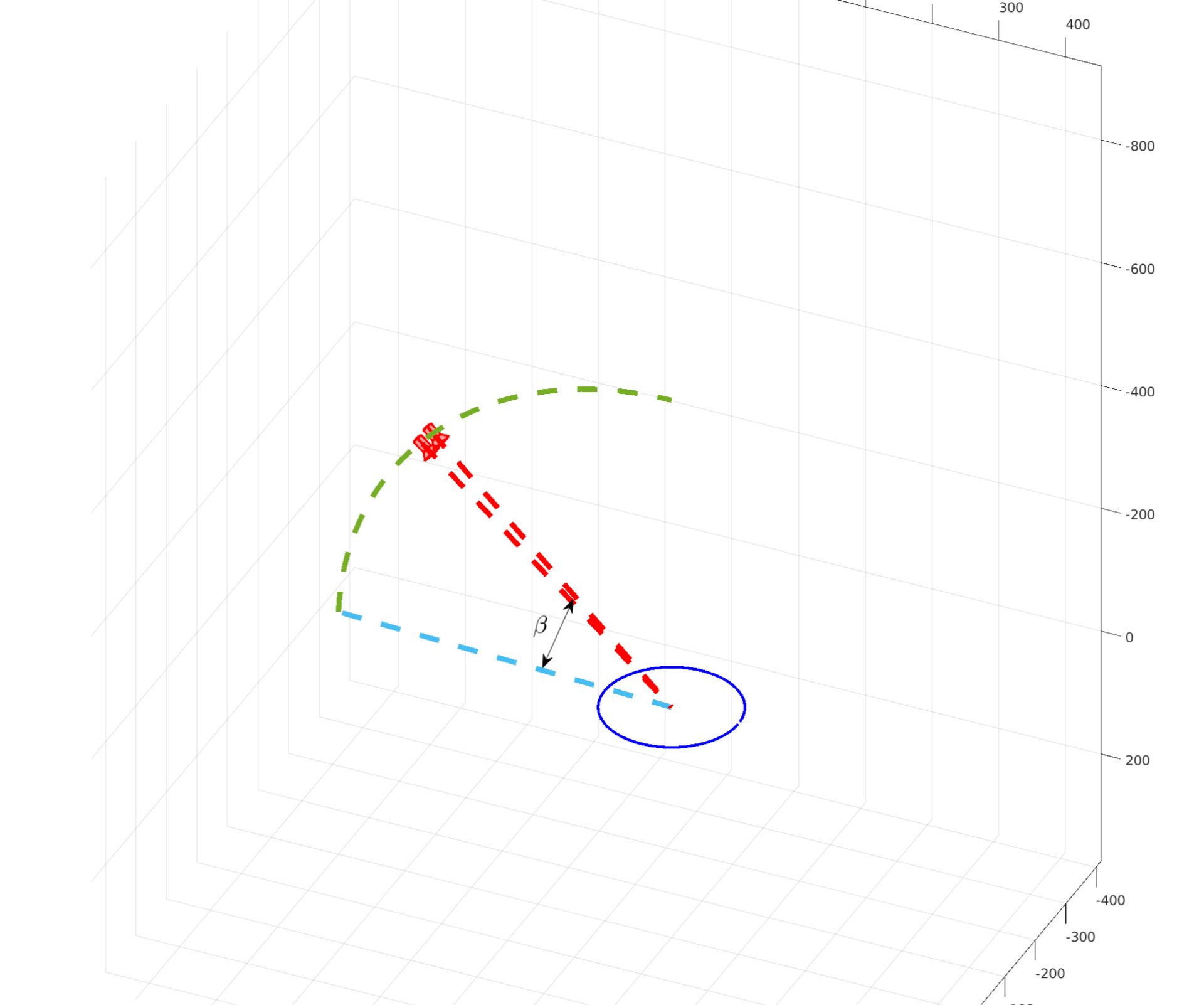}
	    \caption{Incidence angle $\beta$}
     	\label{fig:Angle between ray and vector of direction changing}
	\end{subfigure}
	\hspace{0.1em}
	\begin{subfigure}[ht!]{0.25\textwidth}
	    \includegraphics[scale = 0.23]{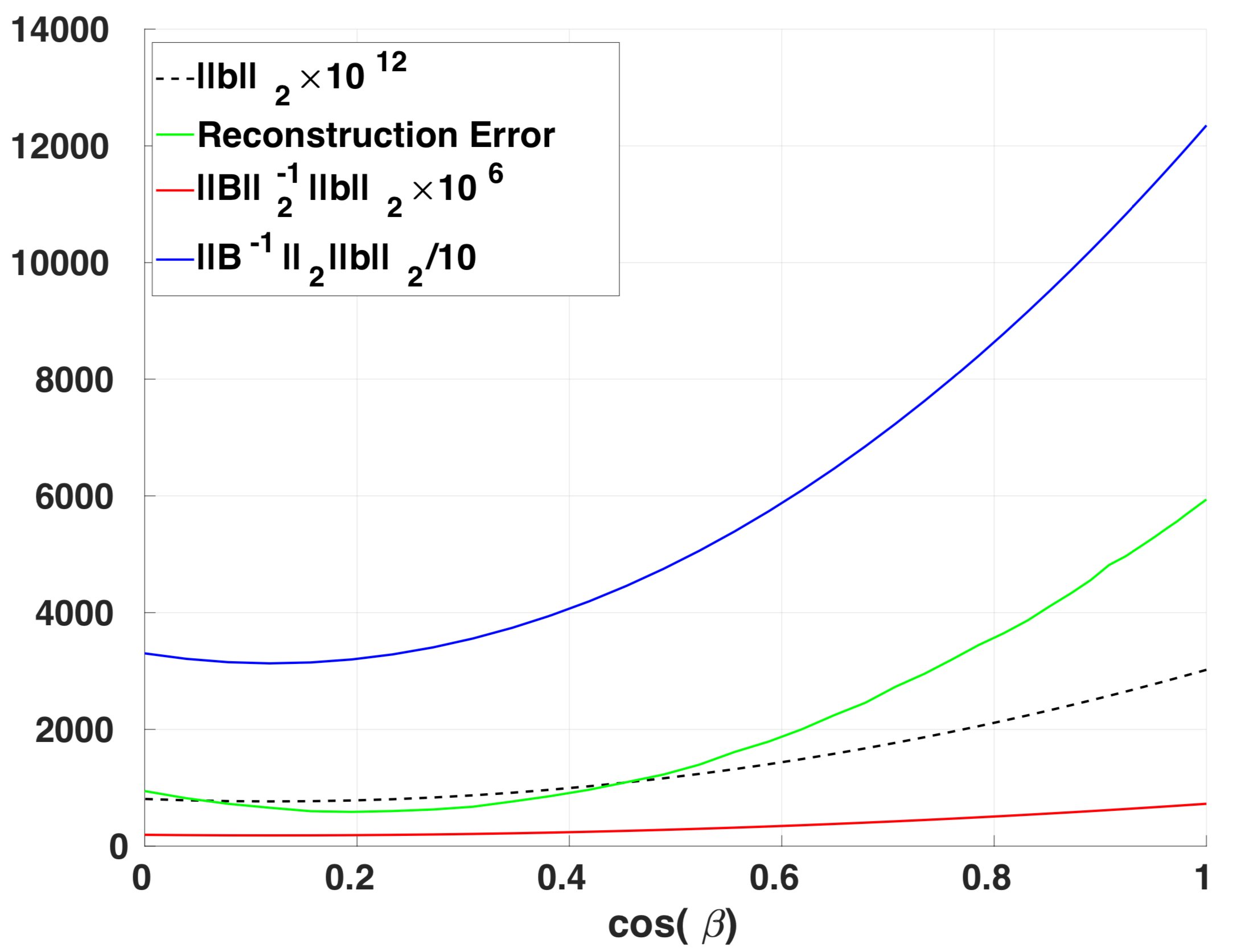}
	    \caption{$\mathbf b$ term analysis}
     	\label{fig:bterm}	    
    \end{subfigure}}
	\caption{
	In (a,b) error bounds specified in Eq. (\ref{ReconBound}) get "tighter" and  reconstruction error is reduced when neighboring viewing rays near orthogonality. In (c,d) as the angle $\beta$ is close to $\pi/2$, both reconstruction error and $\mathbf{||b||}_2$ decrease.}
	
	\label{fig:Term B}
\end{figure*}
We analyze how the Lapalacian linear and quadratic forms  influence the accuracy of our estimates of $\mathbb{X}$, assuming:
%For the  analysis  of our estimate of $\mathbb{X}$ through  Eq. (\ref{LapOptOrigXL}), we assume: 
1) $\mathbb{L}$ is fixed, 2) encodes ground truth temporal adjacency, and 3)  noise free 2D observations.   
%and our optimization has enforced this by assigning a high value to $\lambda_2$. 
This equates to optimizing   Eq. (\ref{LapOptOrigXL}) while  omitting  terms $\mathcal O$ and $\mathcal R$, yielding 
\begin{equation}
    \underset{\mathbb{X}}{\text{min}} \quad  \frac{1}{P}||\mathbb{L} \mathbb{X}||_F^2 +\frac{\lambda_1}{P}tr(\mathbb{X}^\top \LapIO \mathbb{X})
    \label{Eq:AnalyLapQuad}
\end{equation}
We denote the ground truth structure as $\mathbb{X}^*$ and since each point is independently estimated, we analyze the condition of one point per shape.  Then, $\mathbb{X}$  as a point along a viewing ray is
%\begin{equation}
   $ \mathbb{X}_{n,:} = \mathbb{X}^*_{n,:} + l_n \mathbf{r}_n, $
%\end{equation}
where the unknown variables $l_n$ are the signed distance from ground truth along the viewing ray, and $|\mathbf l|$ is the reconstruction error (i.e. depth error).
Eq. (\ref{Eq:AnalyLapQuad}) is an unconstrained quadratic programming problem, solved by setting the derivative over $l$ to zero;  yielding to
\begin{align}
    \mathbb{B}\mathbf{l}  &=  \mathbf{b}
    \label{Eq:DerivaCost} \\
   \mathbb{B}   &=  (\mathbb{L}^{ \top} \mathbb{L}+ \lambda_1 \LapIO) \odot  
    \begin{bmatrix}
        \mathbf{r}_1^\top \mathbf{r}_1 & \dots & \mathbf{r}_N^\top \mathbf{r}_1  \\
        \vdots & \ddots &\vdots \\
        \mathbf{r}_1^\top \mathbf{r}_N & \dots & \mathbf{r}_N^\top \mathbf{r}_N 
    \end{bmatrix} 
    %\mathbb{A}_{f:} = \mathbb{L}^{\mathcal IO}_{f:} diag([r_1^\top r_f,\dots,r_F^\top r_f])
    \label{Eq:B term} \\
    \mathbf b_n  &=  (\mathbb{L}^{\top}_{:,n} \mathbb{L} \mathbb{X}^* + \lambda_1\LapIO_{n,:} \mathbb{X}^*) \mathbf{r}_n
    \label{Eq:b term}
    %(\mathbb{B}^1+\lambda_1\mathbb{B}^2)\mathbf{l} = \mathbf{b}^1+\lambda_2\mathbf{b}^2
\end{align}
where $\mathbb{B}$ is an $N \times N$ matrix and $\mathbf b$ is an $N \times 1$ vector whose $n$-th element is $\mathbf b_n$,
and $\mathbb{L}_{:,n}$ denotes the $n$-th column of $\mathbb{L}$ and $\LapIO_{n,:}$ denotes the $n$-th row. From Eq. (\ref{Eq:DerivaCost}), we attain the lower and upper bounds for reconstruction error as
\begin{equation}
    ||\mathbb{B}||_2^{-1}||\mathbf{b}||_2 \leq ||\mathbf{l}||_2  \leq ||\mathbb{B}^{-1}||_2||\mathbf{b}||_2 
    \label{ReconBound}
\end{equation}

%------------------------------------------------------------------------
\noindent {\bf Imaging geometry convergence.}
We consider two cameras alternating the 
%temporal %
capture of a motion sequence, which  are placed sufficiently far 
%To show how this influences reconstruction accuracy, we set two cameras far away 
from the motion center $\mathbf{c}$, such that the viewing ray convergence angle for all joints can be approximated by the angle $\theta$ between the cameras to the motion center.  We vary $\theta$ from 0 to $\pi$ as in Fig. \ref{fig:Angle between cameras}.and evaluate the reconstruction error and upper bounds, which as shown in Fig. \ref{fig:Bterm}  decrease as viewing rays approach orthogonality. 

%------------------------------------------------------------------------
\noindent{\bf 3D motion observability.} 
The vector $\LapIO_{n,:} \mathbb{X}^*$ in Eq. (\ref{Eq:b term}),  lies on a local motion plane  formed by $\mathbb{X}^*_{n,:}$ and it's two neighboring points. Similarly,each row in $\mathbb{L} \mathbb{X}^*$ will also be a vector on a  local motion plane. For %sufficiently 
smooth motion under dense sampling, a triplet of successive local motion planes can be approximated by a common 3D plane $\pi_n$. Hence,  the vector $\mathbb{L}^{\top}_{:,n} \mathbb{L} \mathbb{X}^* + \lambda_1\LapIO_{n,:} \mathbb{X}^* $ will be contained in $\pi_n$, yielding  smaller values of $\mathbf b_n$ as $\pi_n$ and the viewing rays $\mathbf r_n$ near orthogonality.In Fig. \ref{fig:Angle between ray and vector of direction changing}, we consider a circle motion observed by two cameras with constant convergence angle, pointing to the motion center. In this configuration,  $||\mathbb{B}||_2^{-1}$ and $||\mathbb{B}^{-1}||_2$ are nearly constant. We vary the angle $\beta$ between the viewing directions and the motion plane $\pi_n$. Fig. \ref{fig:bterm} shows more accurate reconstruction is attained for viewing directions near orthogonal to the motion plane.

%------------------------------------------------------------------------
\section{Experiments}
\begin{figure*}[t!] 
	\begin{subfigure}[ht!] {0.6\textwidth} % width of left 
		\includegraphics[scale = 0.775]{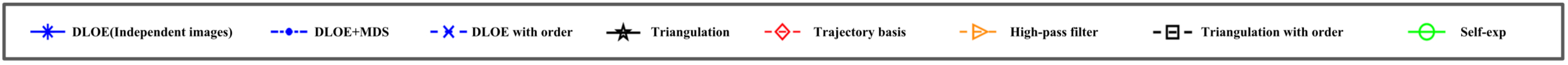}
	\end{subfigure}
	
	\fbox{\hspace{0.3cm}\begin{subfigure}[ht!] {0.245\textwidth} % width of left 
		\includegraphics[scale = 0.23]{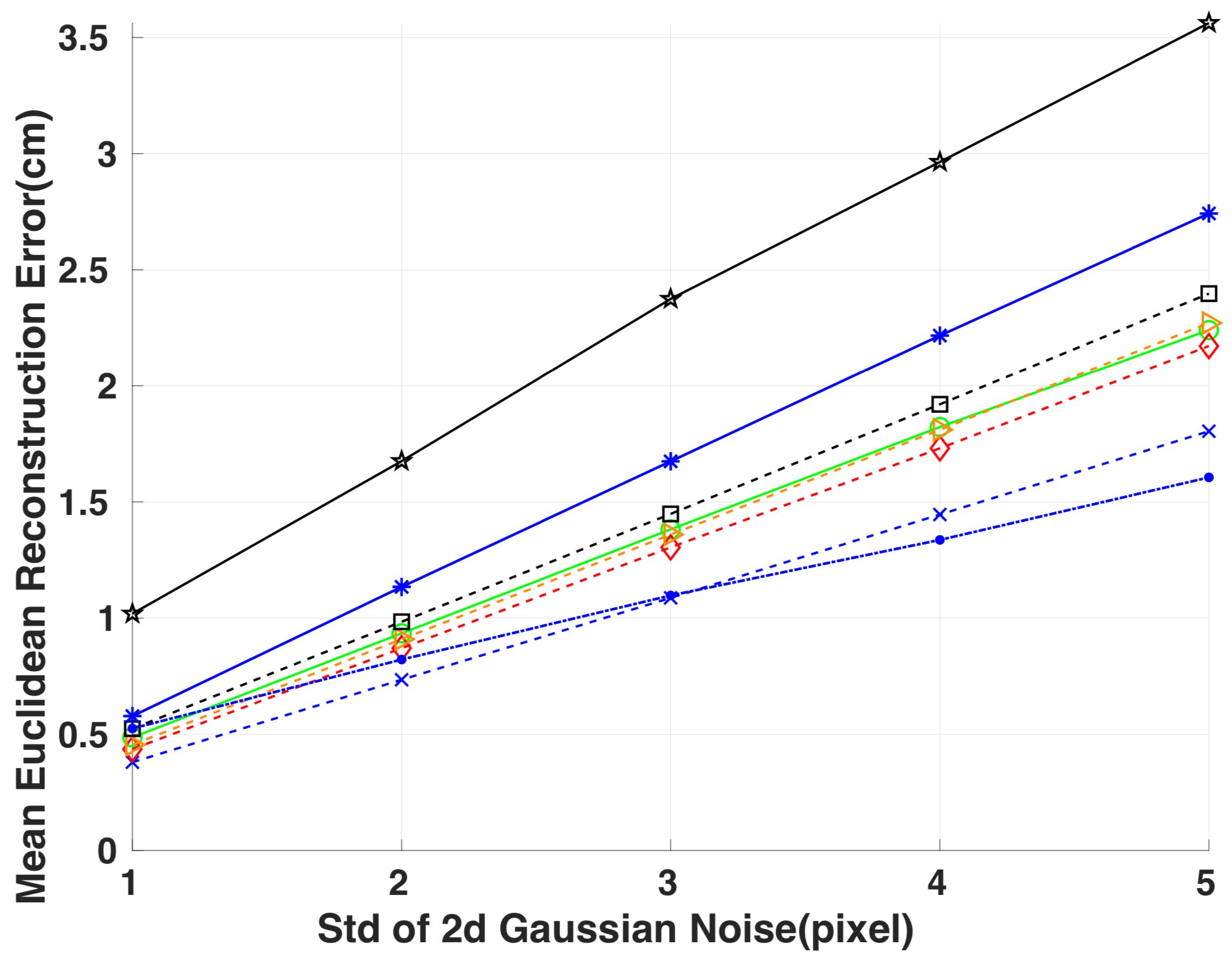}
		\caption{Noise level} 
		\label{fig:DiffNoise}
	\end{subfigure}
	%\hspace{1em}
	\begin{subfigure}[ht!] {0.245\textwidth} % width of right 
		\includegraphics[scale = 0.23]{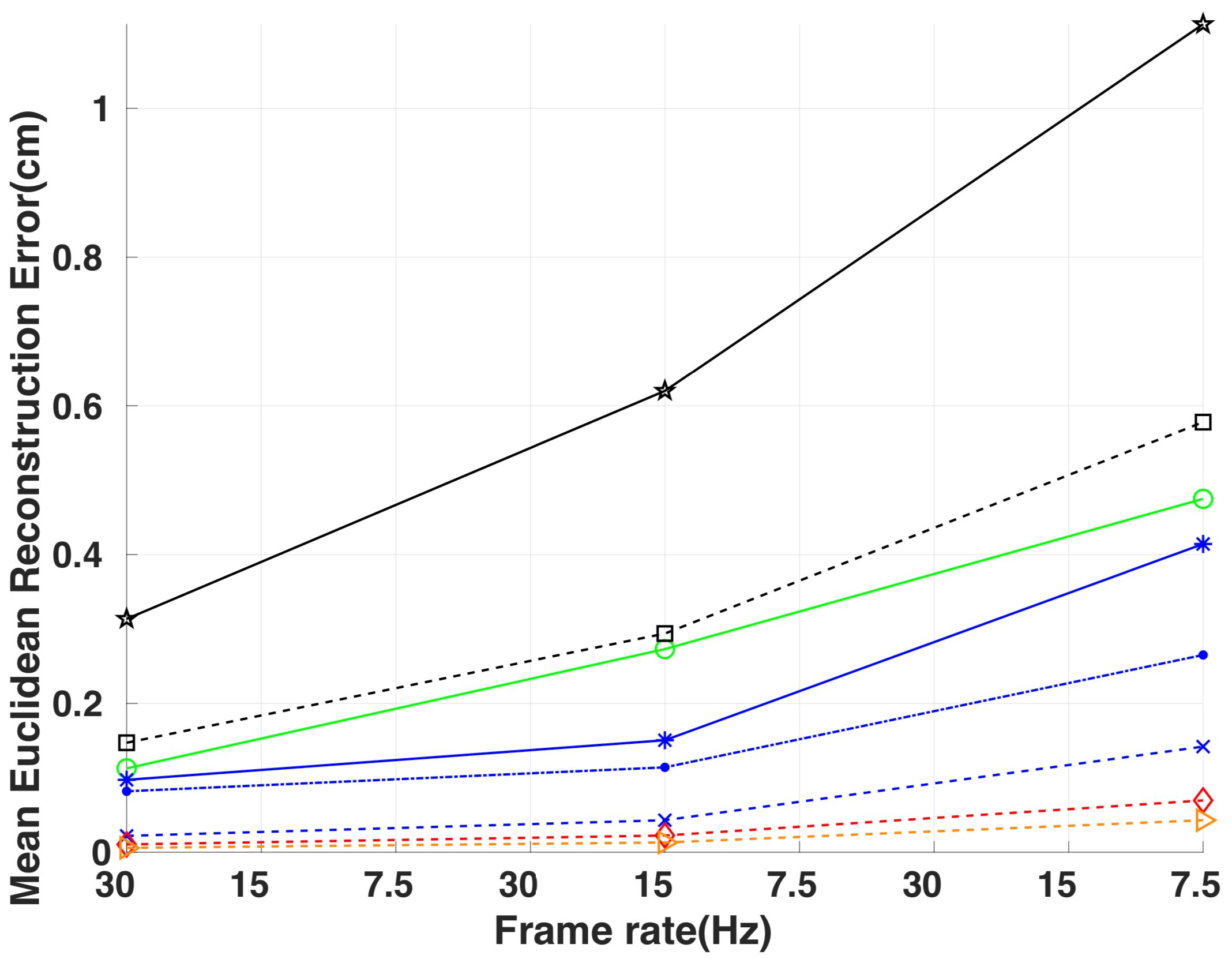}
		\caption{Frame rates} 
		\label{fig:DiffFreq}
	\end{subfigure}
	%\hspace{1em}
	\begin{subfigure}[ht!] {0.245\textwidth} % width of right 
		\includegraphics[scale = 0.23]{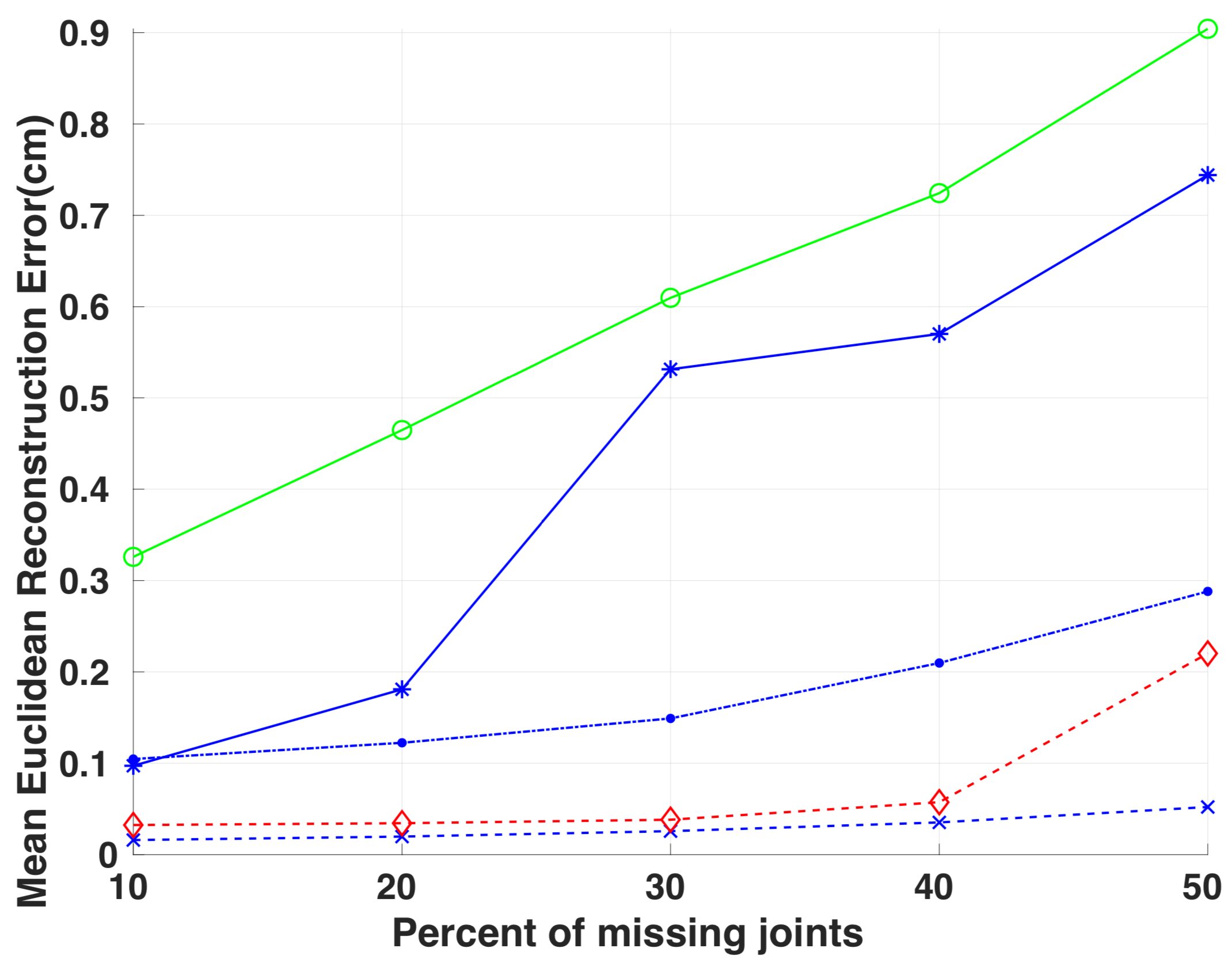}
		\caption{Missing points} 
		\label{fig:DiffMissing}
	\end{subfigure}
	%\hspace{1em}
	\begin{subfigure}[ht!] {0.245\textwidth} % width of right 
		\includegraphics[scale = 0.23]{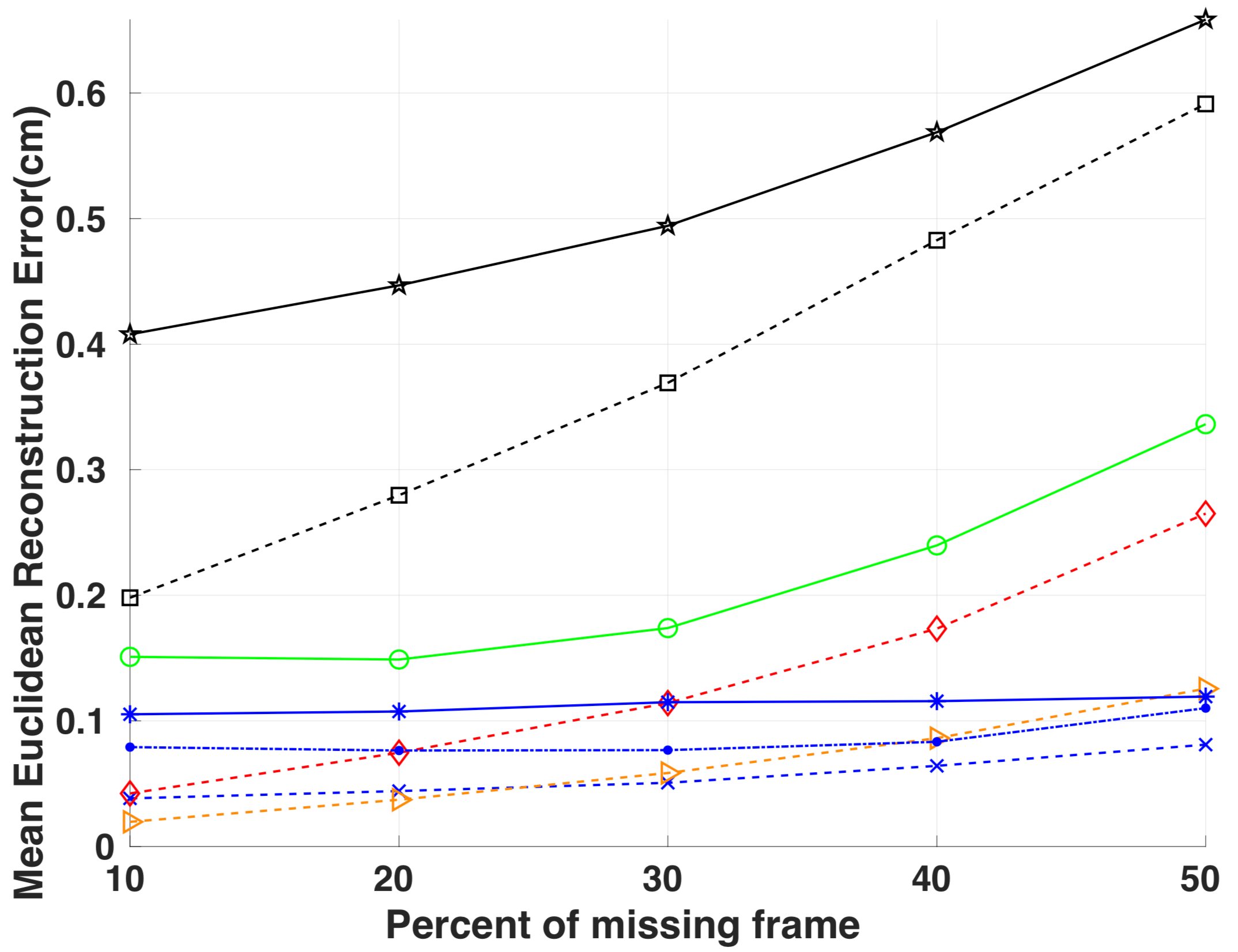}
		\caption{Nonuniform density} 
		\label{fig:DiffFrameDrop}
 	\end{subfigure}}
	\caption{Reconstruction error for motion capture data under different conditions. Reported averages over 20 executions.} % 
\end{figure*}

\begin{figure*}[ht!]
    \fbox{\begin{subfigure}[ht!] {0.3\textwidth} % width of right 
        \centering
        \includegraphics[scale = 0.06, trim={7cm 7cm 7cm 7cm},clip]{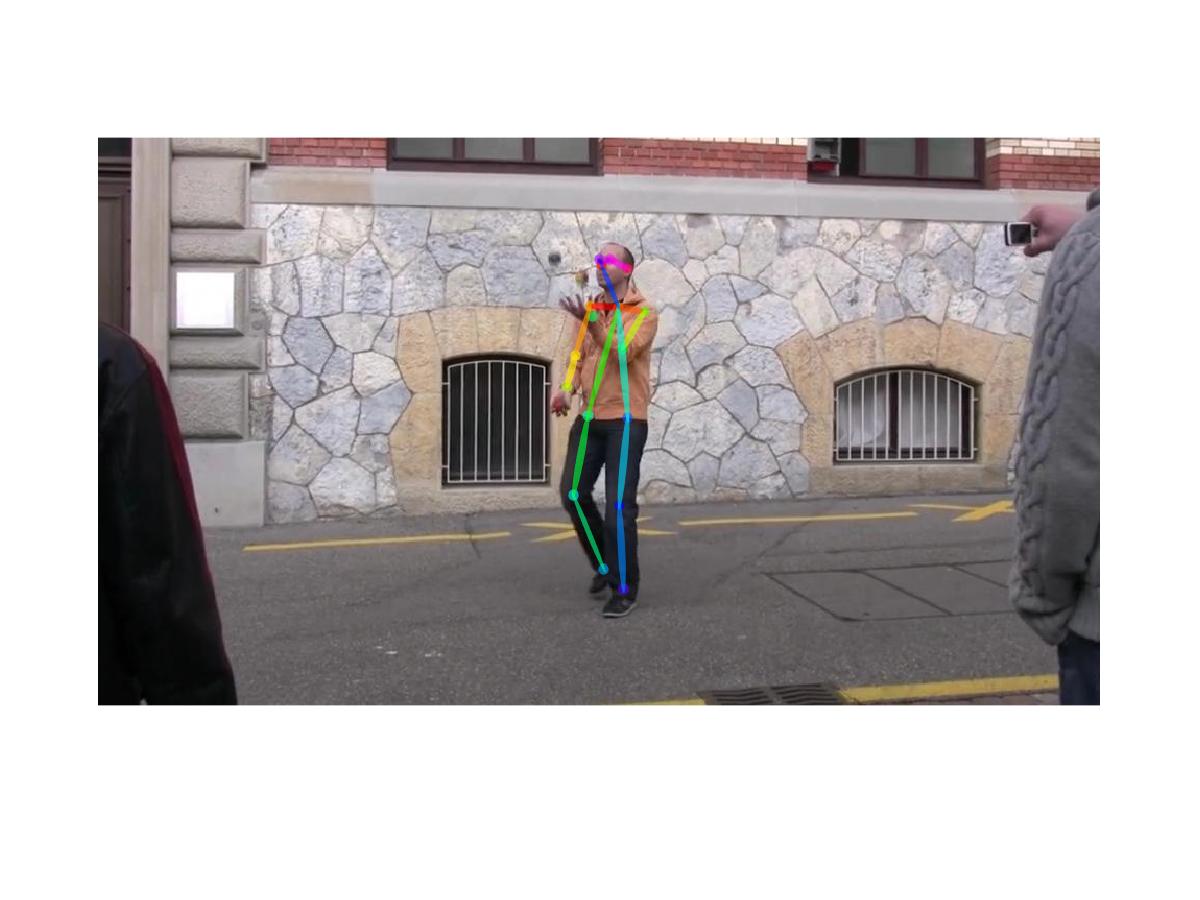} \quad
        \includegraphics[scale = 0.13]{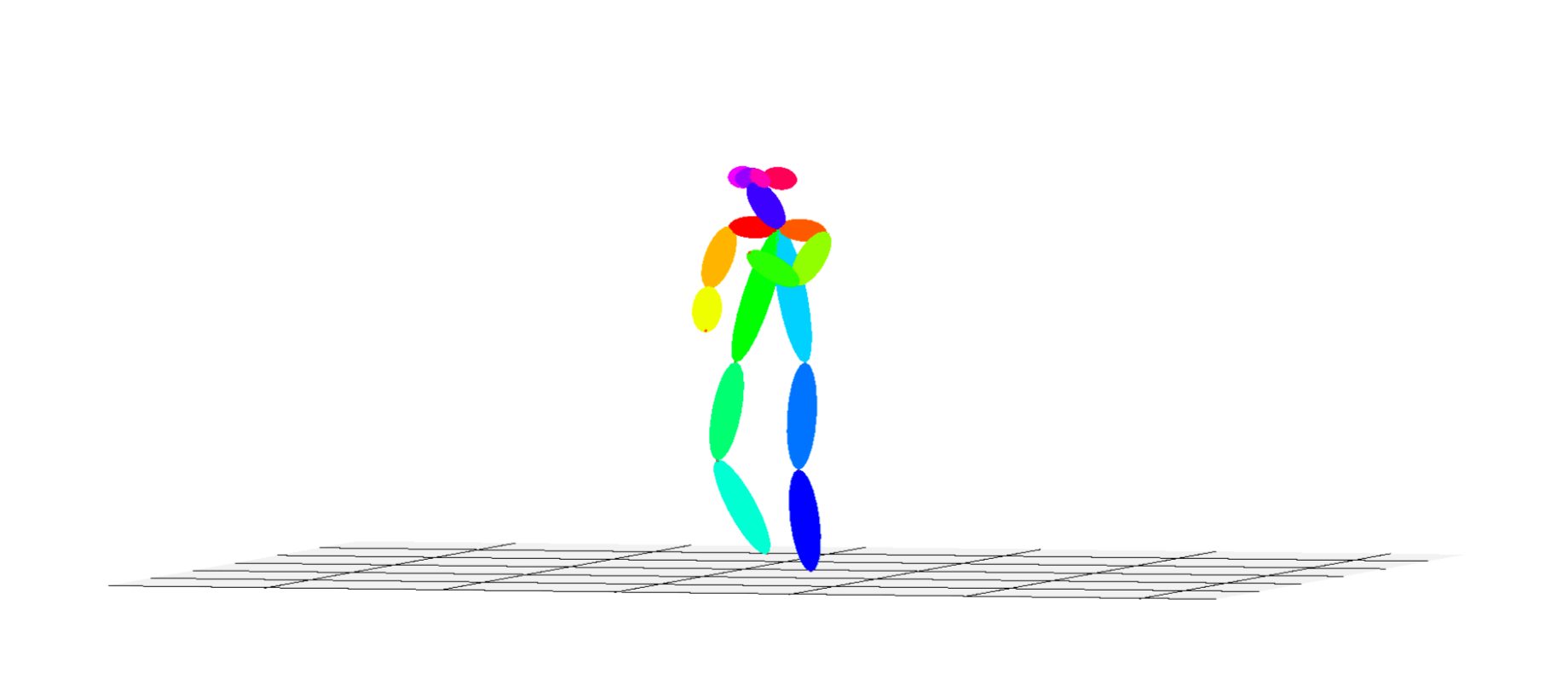} \quad
        
        \includegraphics[scale = 0.06, trim={7cm 7cm 7cm 7cm},clip]{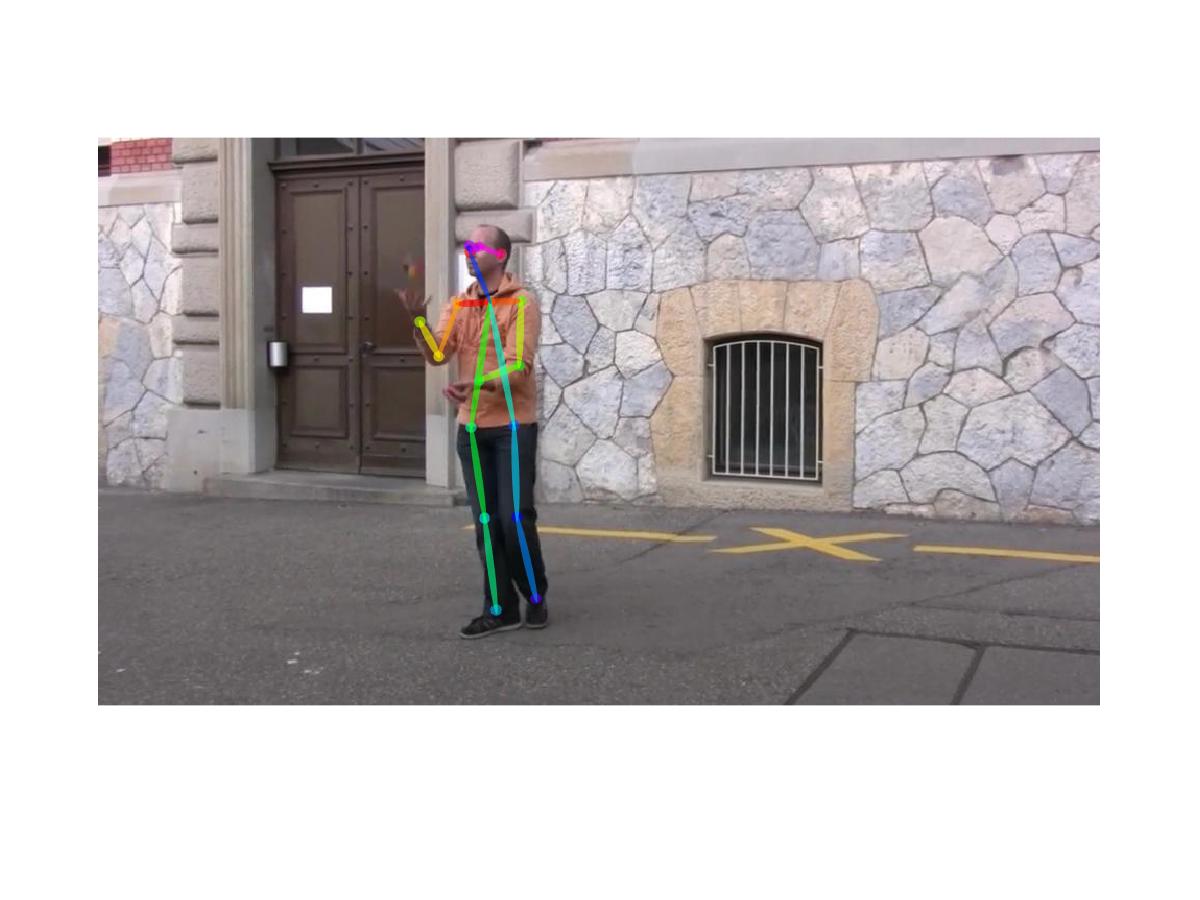} \quad
        \includegraphics[scale = 0.13]{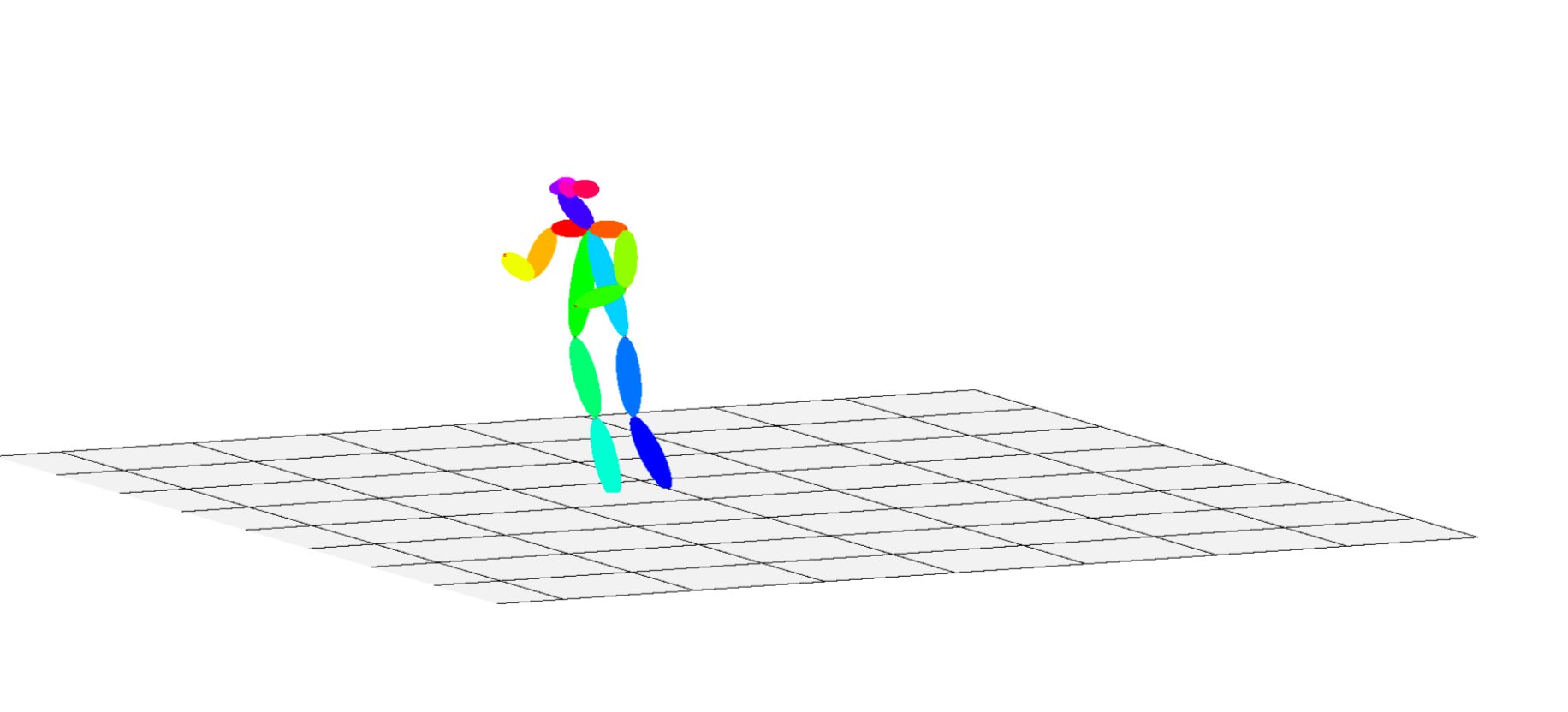} \quad
        
        \includegraphics[scale = 0.06, trim={7cm 7.8cm 7cm 6.2cm},clip]{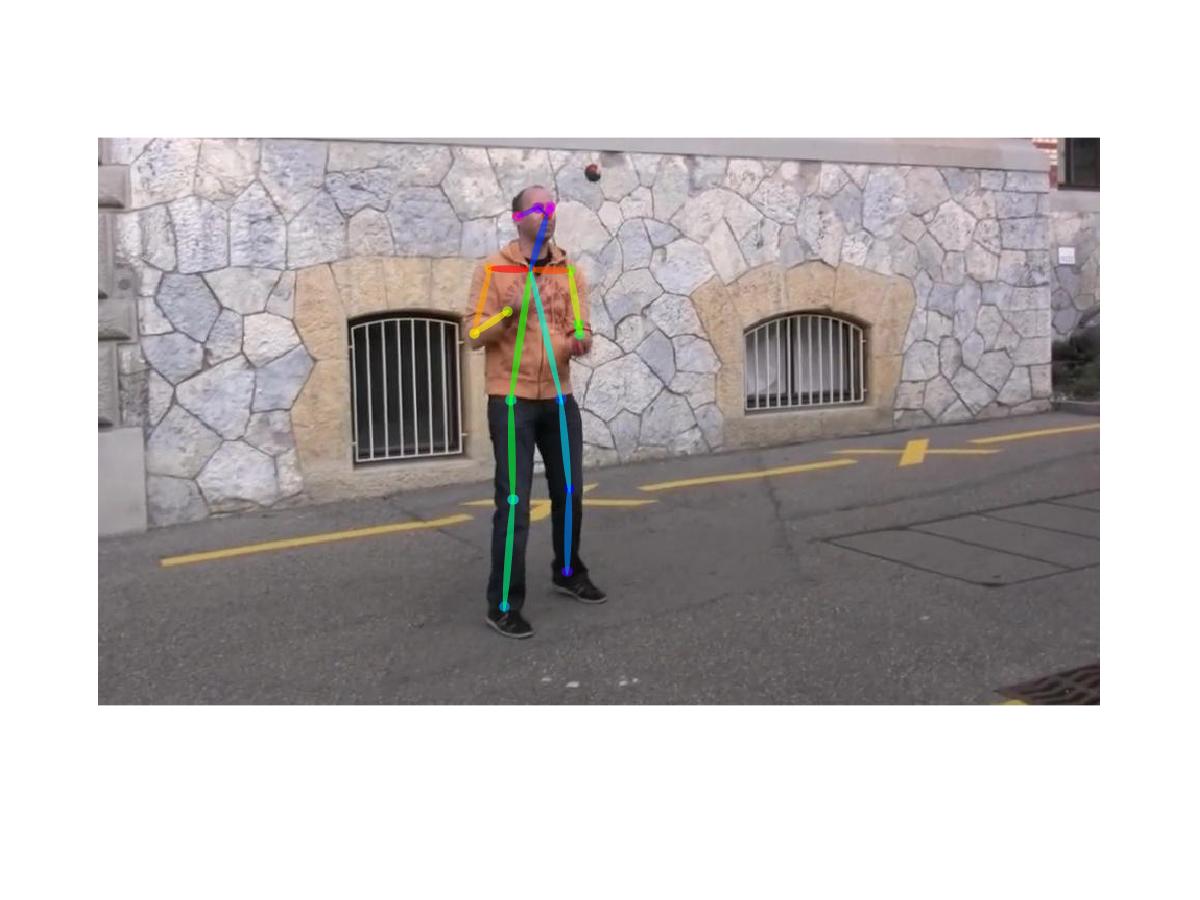} \quad
        \includegraphics[scale = 0.12]{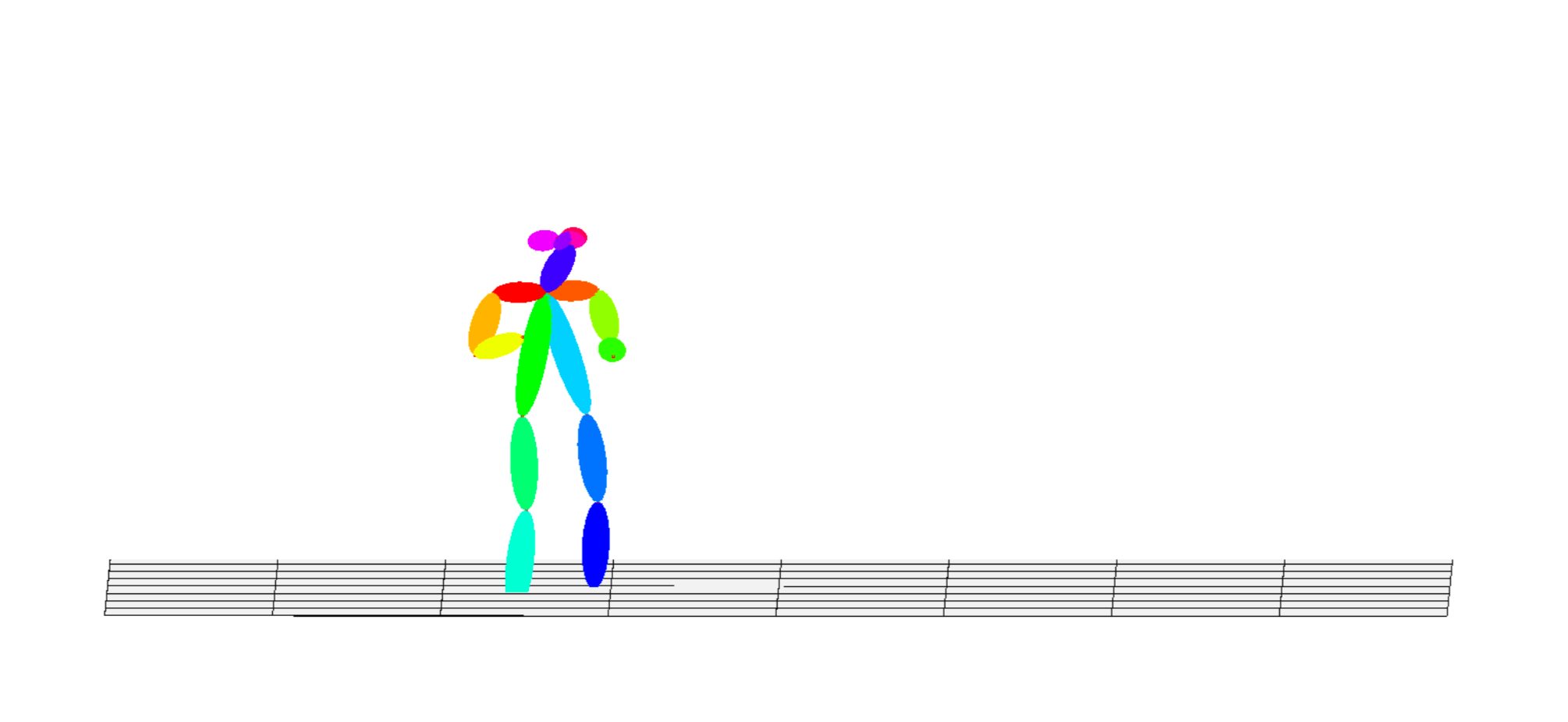} \quad
        
         \includegraphics[scale = 0.06, trim={7cm 7.8cm 7cm 6.2cm},clip]{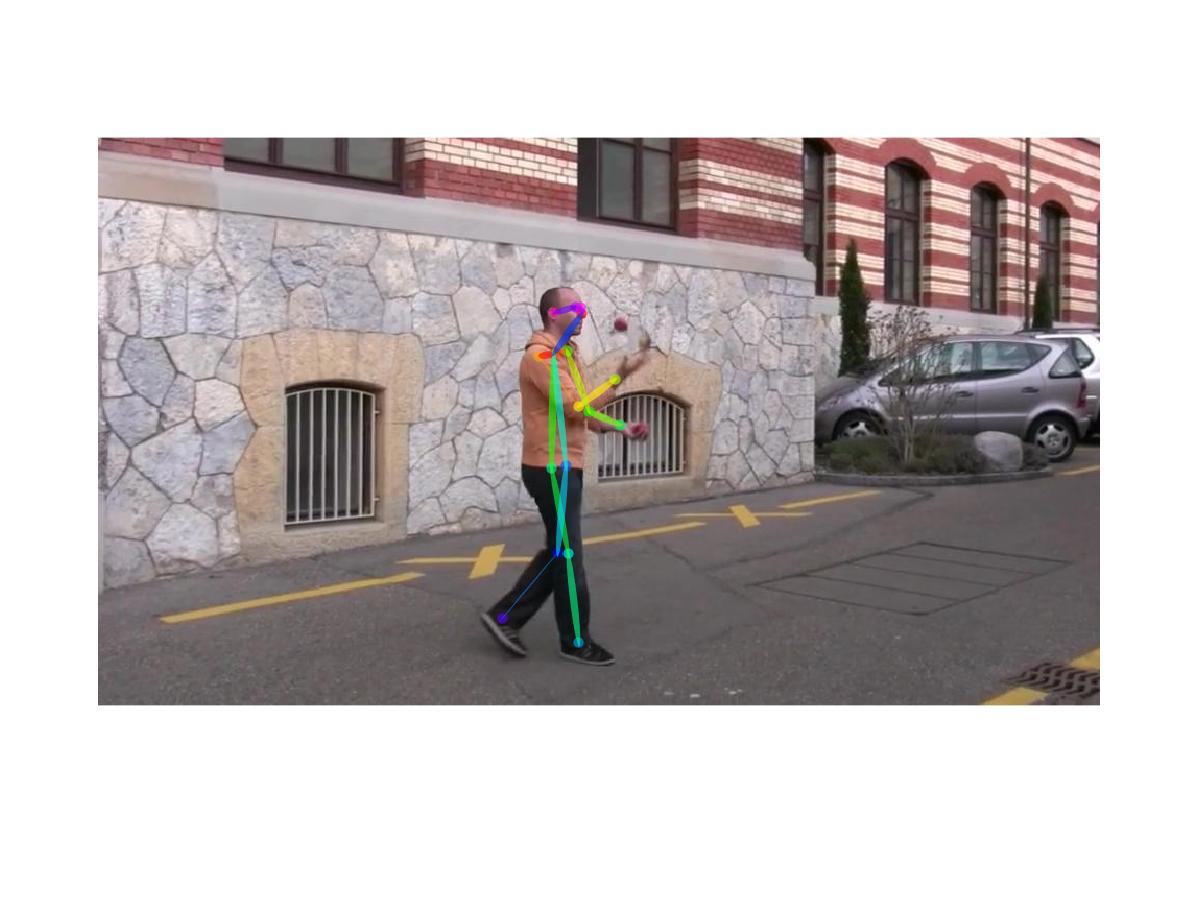} \quad
        \includegraphics[scale = 0.138]{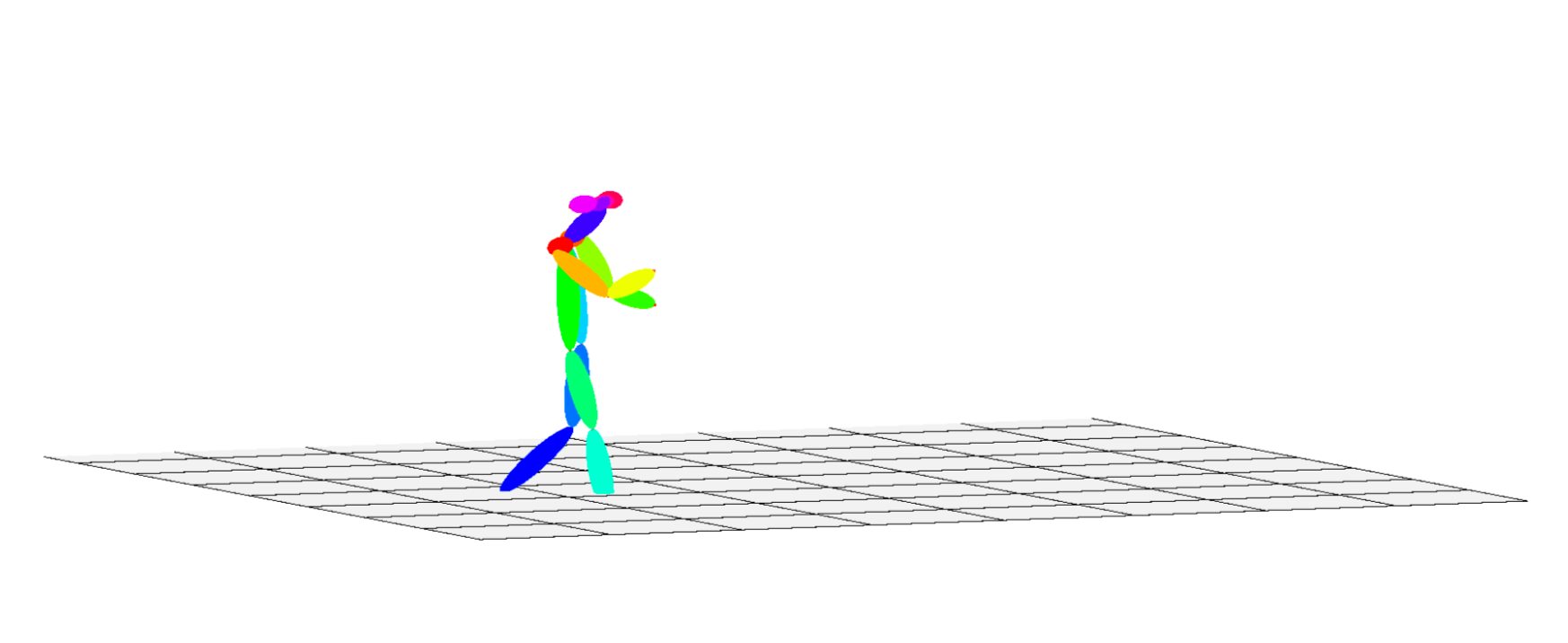} \quad
        
        \caption{ Juggler.} % caption for whole figure
    	\label{fig:Juggler}
    \end{subfigure}}
    \fbox{\begin{subfigure}[ht!]{0.4\textwidth} % width of right 
        \centering
        \includegraphics[scale = 0.368, resolution = 72]{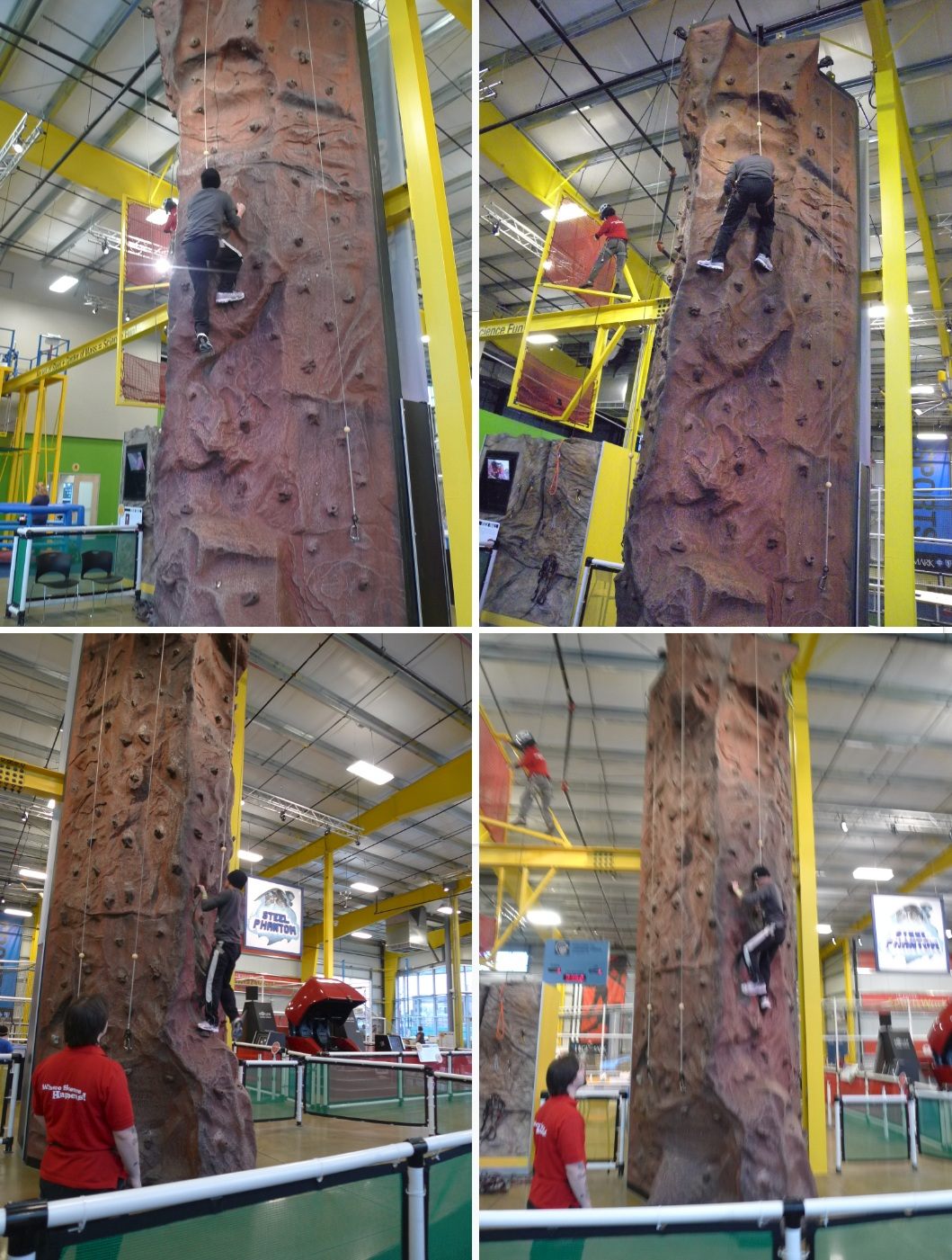} \quad
        \includegraphics[scale = 0.48]{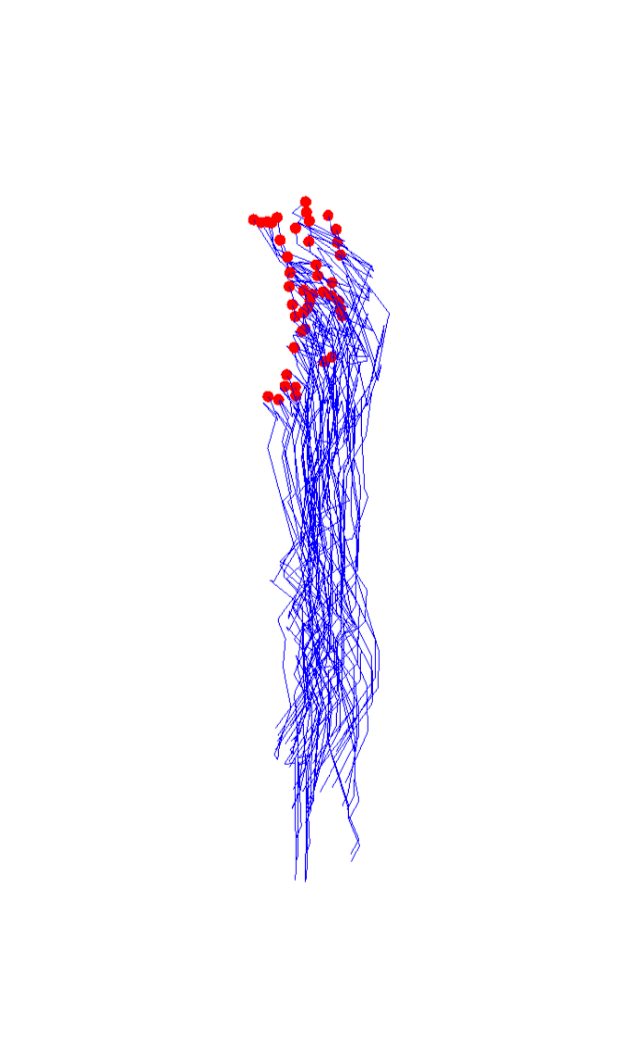}
        \caption{Climb}
    	\label{fig:Climb}
	\end{subfigure}}
    \fbox{\begin{subfigure}[ht!]{0.3\textwidth} % width of right 
        \centering
        \includegraphics[scale = 0.295, resolution = 72]{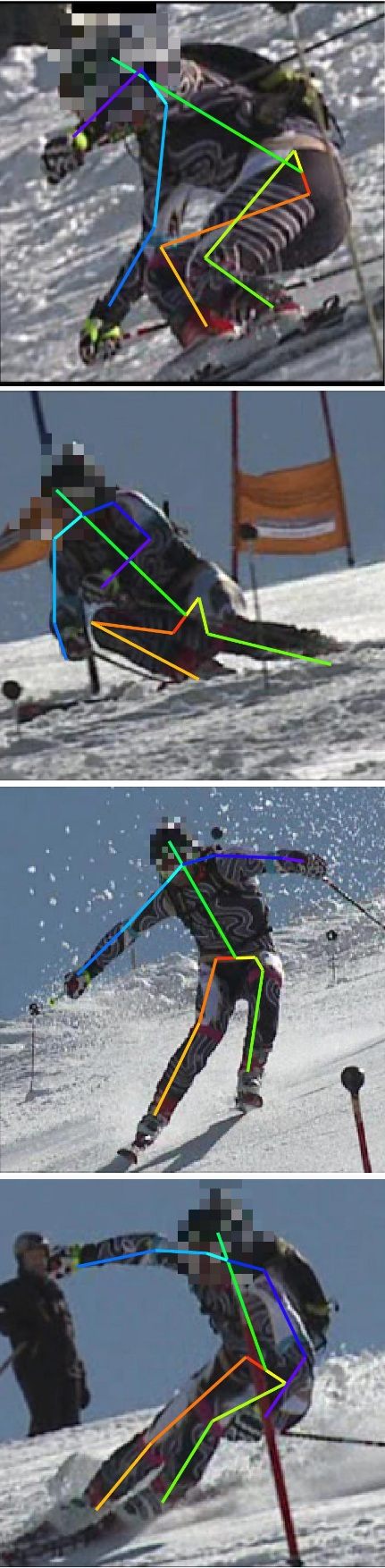}
        \includegraphics[scale = 0.345]{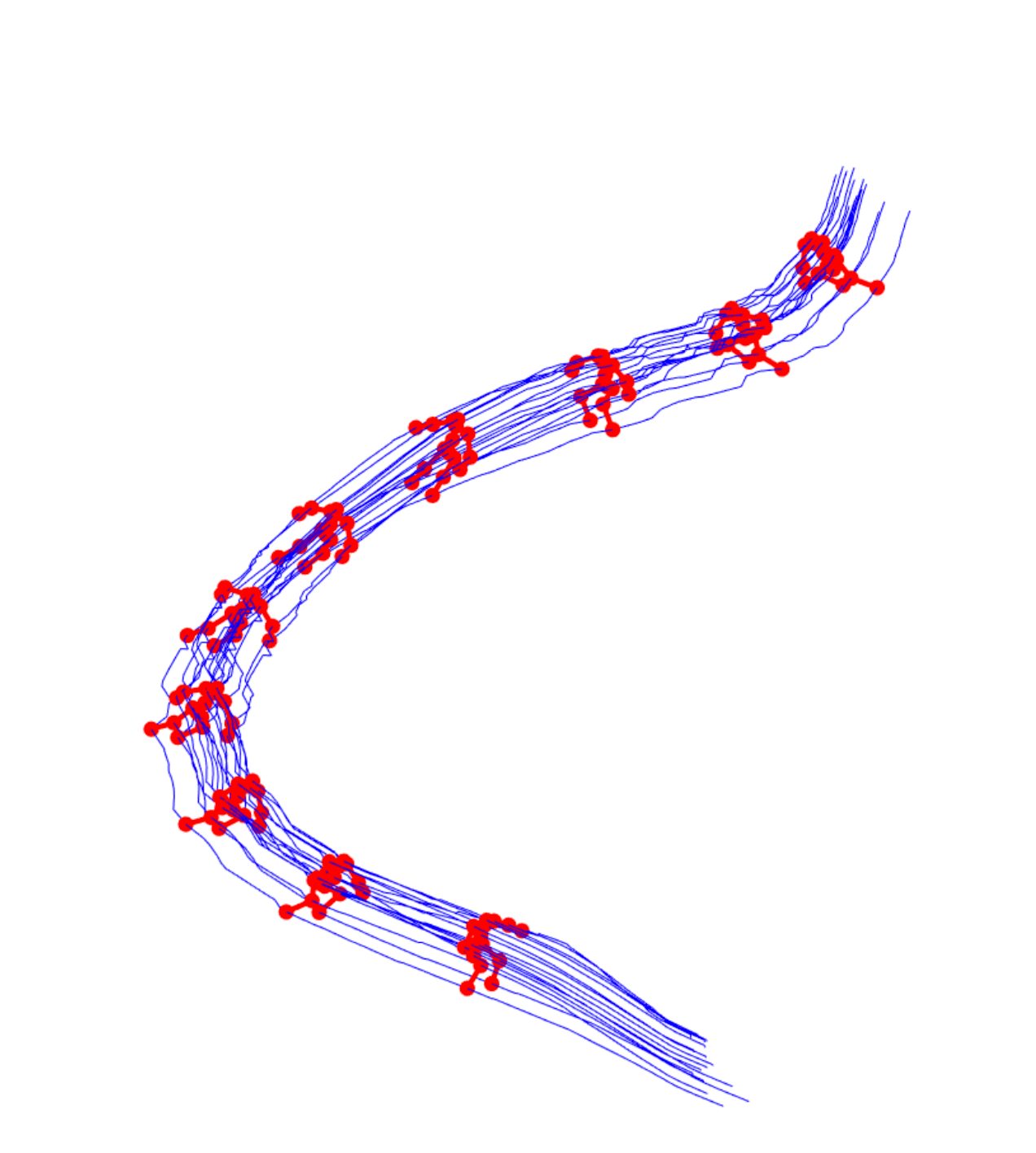}
        \caption{Ski}
    	\label{fig:Ski}
	\end{subfigure}}
	
	\begin{center}
    	\begin{subfigure}[ht!]{0.8\textwidth} % width of right 
            \resizebox{\columnwidth}{!}{%
                \centering
                \begin{tabular}{|c|c|c|c|c|c|c|c|c|}
                \hline
                &data type & motion type & solver type & number of cameras & number of frames & number of joints & frame rate & kendall rank correlation \\
                \hline
                Juggler & unsynchronized videos & repeating motion & DLOE+MDS+$\mathbb{W}_{prior}$  & 4 & 80 & 18 & 6.25 & 0.8816 \\
                \hline
                Climb & unsynchronized images & linear motion & DLOE+MDS+$\mathbb{W}_{prior}$ & 5 & 27 & 45 & N/A & 0.8689\\
                \hline
                Ski & unsynchronized videos & nonlinear motion & DLOE+MDS+$\mathbb{W}_{prior}$ & 6 & 137 & 17 & N/A & 0.9526\\
                \hline
                \end{tabular}
            }
            %\caption{Datasets information}
    	\end{subfigure}
	\end{center}
	
	\caption{Experiments on multi-view image capture. All datasets were devoid of concurrent observations. }
	\label{fig_MultiViewDatasets}
\end{figure*}

\begin{figure*}[ht!]
    \centering
    \fbox{\begin{subfigure}[ht!]{0.52\textwidth} % width of right 
        \begin{minipage}[b]{0.431\linewidth}
            \includegraphics[scale = 0.57, resolution = 72]{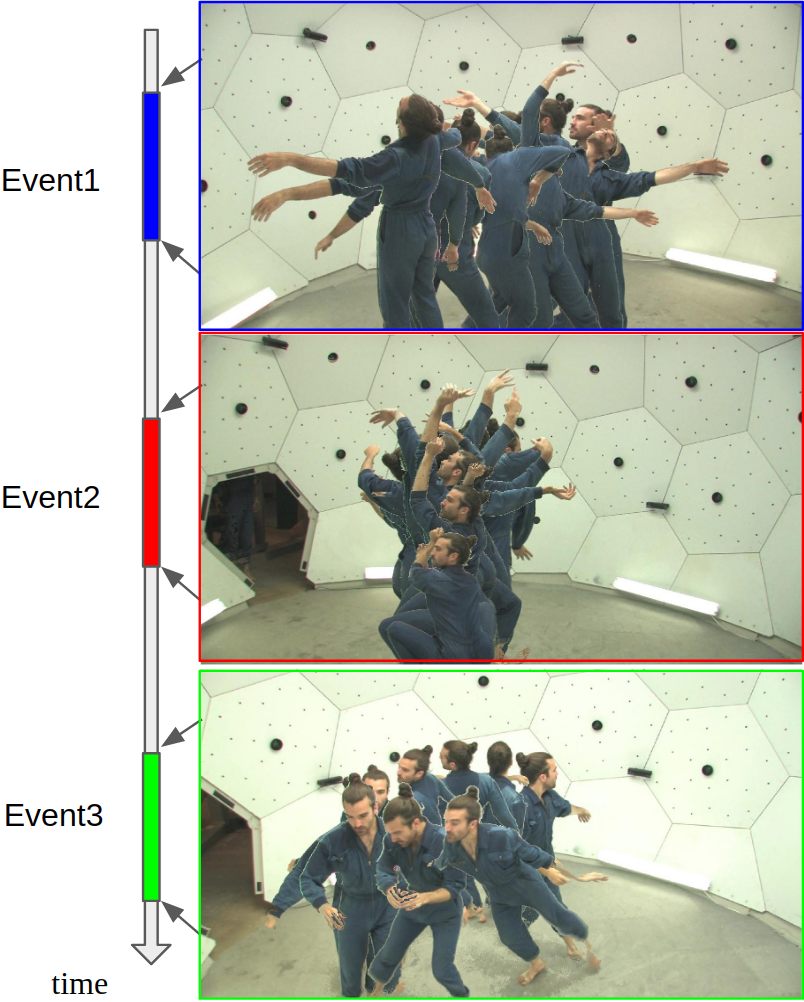}
        \end{minipage}
        \qquad
        \begin{minipage}[b]{0.2\linewidth}
            \includegraphics[scale = 0.082, resolution = 72]{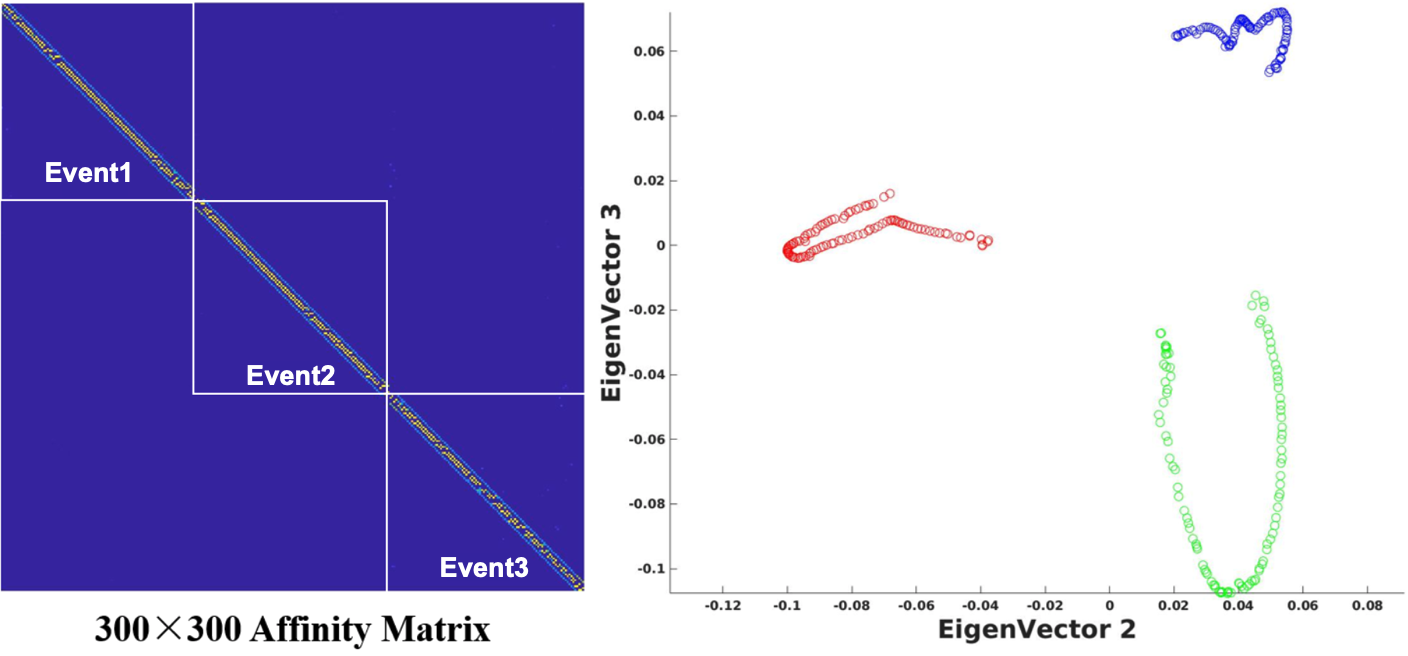}

            \includegraphics[scale = 0.3]{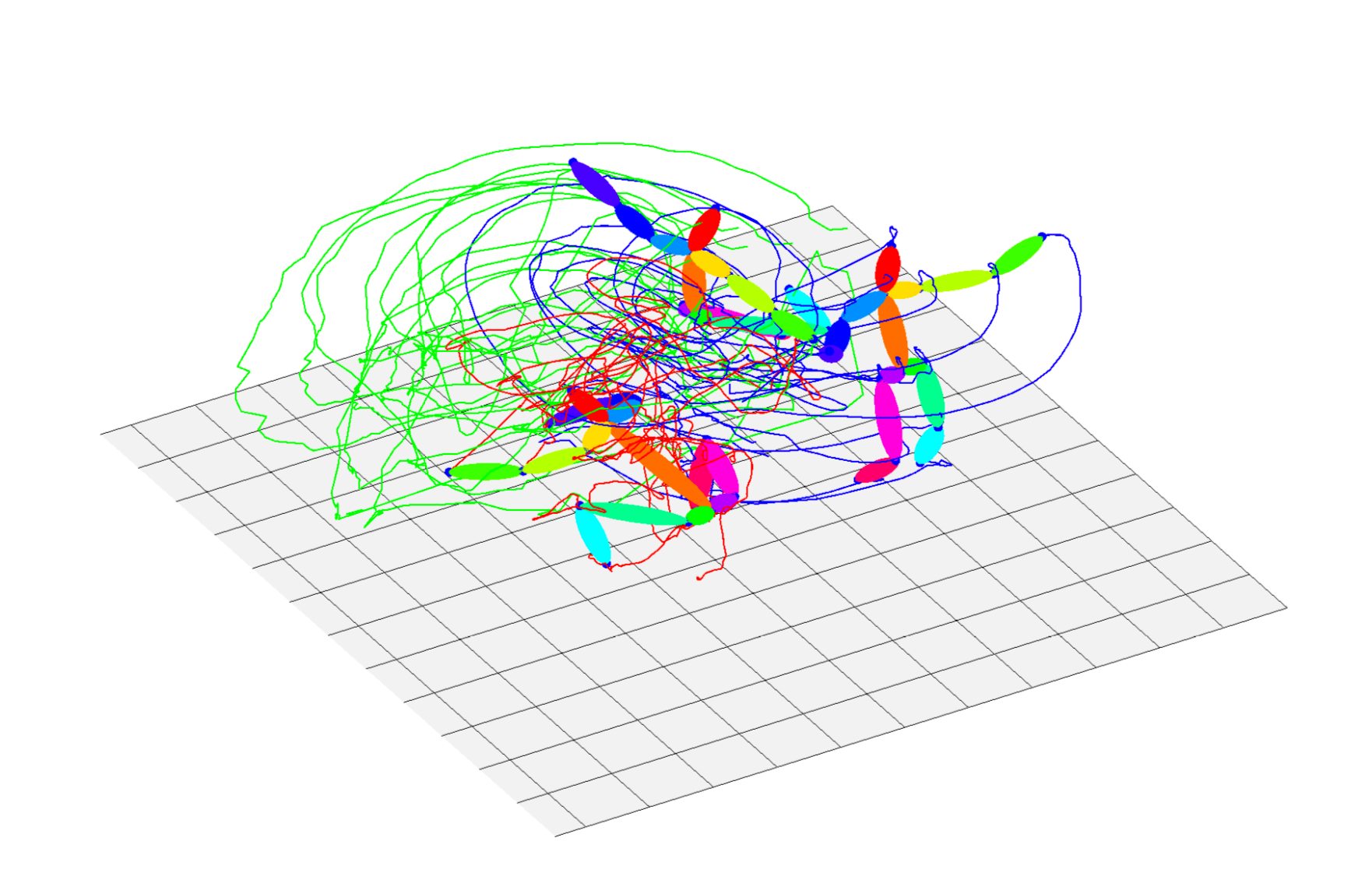}
        \end{minipage}
        \caption{Event segmentation \label{fig_EventSegmentation}}
    \end{subfigure}}
    \hspace{0.15cm}
    \fbox{\begin{subfigure}[ht!]{0.45\textwidth} % width of right 
        \begin{minipage}[b]{0.4\linewidth}
            \centering
            \includegraphics[scale = 0.29, resolution = 72]{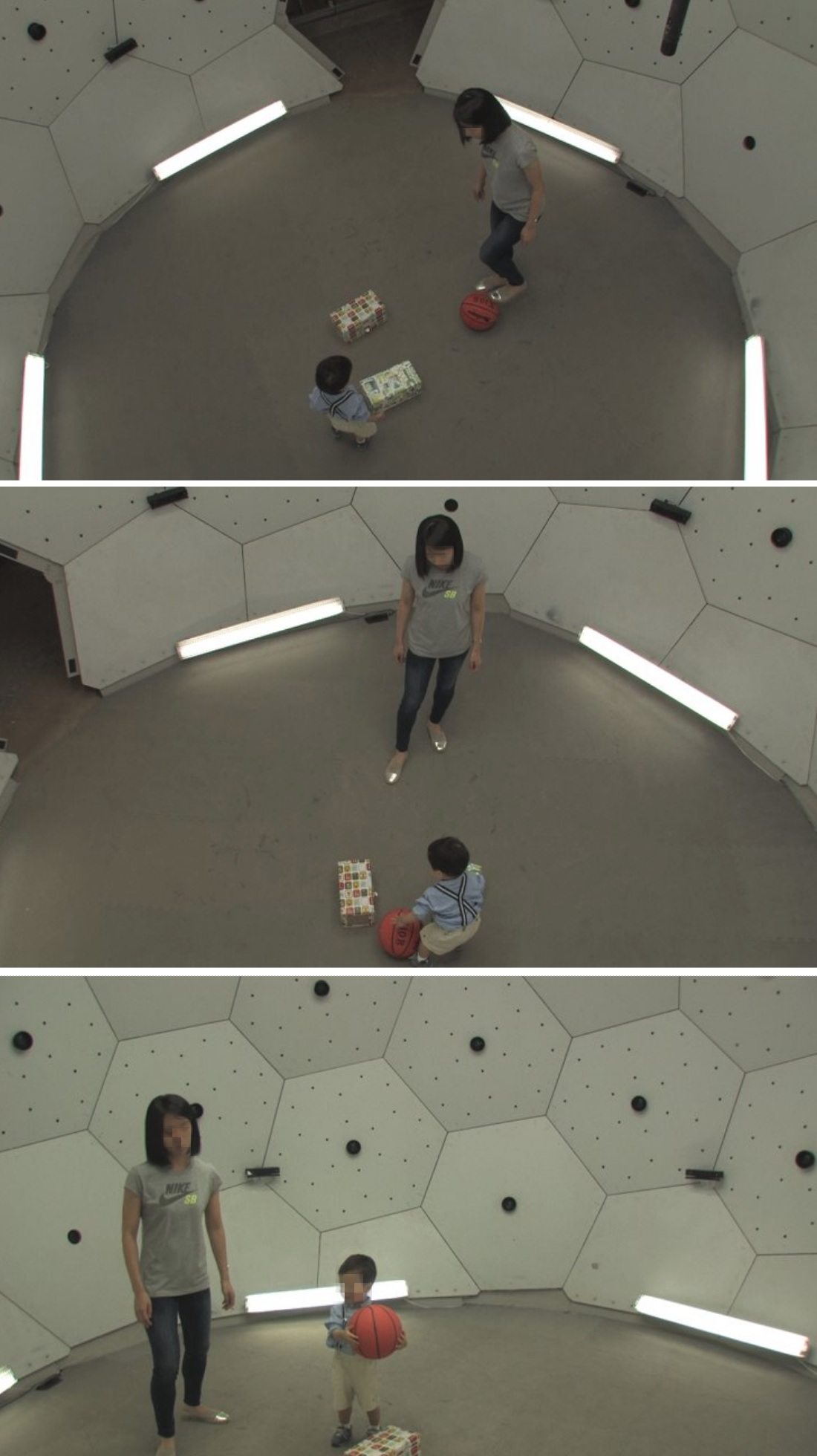}
        \end{minipage}    
        \quad
        \begin{minipage}[b]{0.1\linewidth}
            \centering
            \includegraphics[scale = 0.075, resolution = 72]{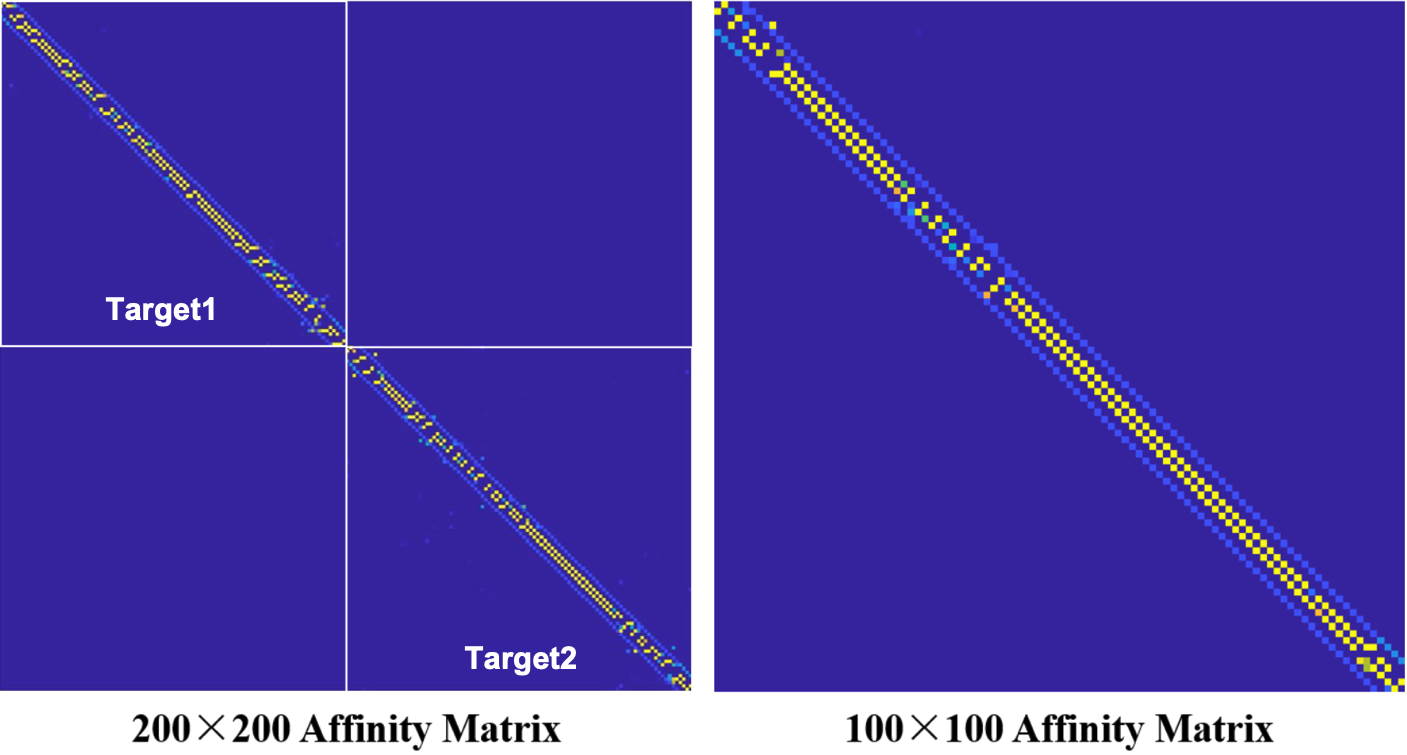}
            \includegraphics[scale = 0.3]{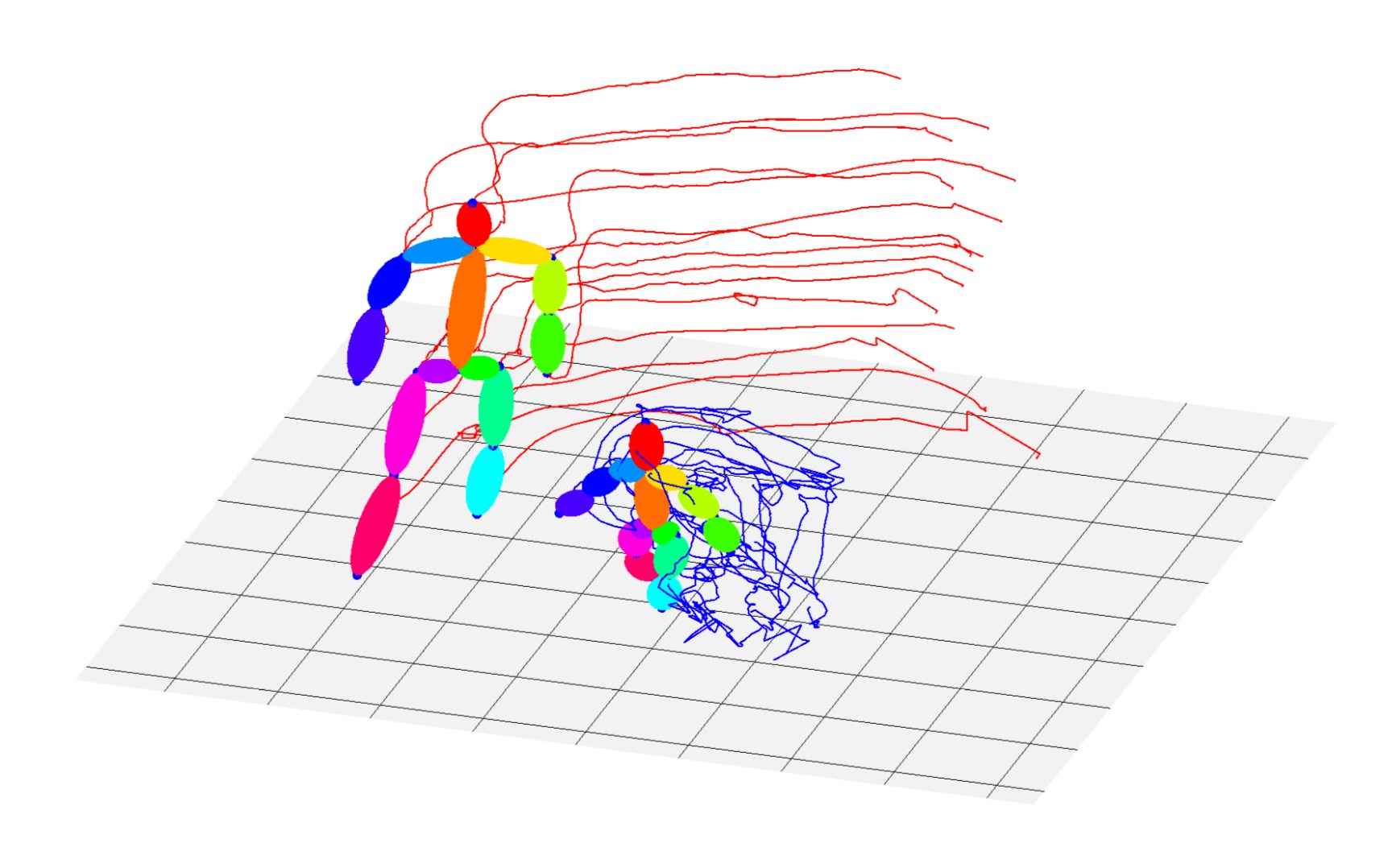}
        \end{minipage}  
        \caption{Multi-Target scenario \label{fig_MultiPerson} }
	\end{subfigure}}
	
	\begin{center}
    	\begin{subfigure}[ht!]{0.8\textwidth} % width of right 
            \resizebox{\columnwidth}{!}{%
                \centering
                \begin{tabular}{|c|c|c|c|c|c|c|c|c|}
                \hline
                &data type & motion type & solver type & number of cameras & number of frames & number of joints & frame rate & kendall rank correlation \\
                \hline
                dance & unsynchronized videos & nonlinear motion & DLOE+$\mathbb{W}_{prior}$ & 4 & 300 & 15 & 7.5 & 0.9802 \\
                \hline
                multi-person & independent images & nonlinear motion& DLOE+$\mathbb{W}_{prior}$ & 4 & 100 & 15 & 7.5 & 1\\
                \hline
                \end{tabular}
            }
            %\caption{Datasets information}
    	\end{subfigure}
	\end{center}
	\caption{Results on Dancing and Toddler \cite{joo2017panoptic}.  Disjoint Dancing segments form an input datum. Spectral visualization of  estimated affinity matrix reveal a triplet of clusters. For Toddler, we use DLOE for instance identification, see text for details.}
\end{figure*}

\subsection{Motion Capture Datasets} \label{Artificial} 
We synthesize 2D features of human 3D motions for 31 joints with frame rates of 120 Hz \cite{muller2007documentation}. We choose 10 sample motions, each having on average $\sim$300 frames. We use the 3D joint positions as ground truth dynamic structure and project them to each frame on four virtual cameras as 2D observations. All cameras  have $1000 \times 1000$ resolution and 1000 focal length, are static with a distance   of 3 meters around the motion center. The four {\bf cameras are unsynchronized}, with frame rate up to 30 Hz. Accuracy is quantified by mean 3D reconstruction error. 
Our method discrete Laplace operator estimation (DLOE) is compared against self-expressive dictionary learning (SEDL)\cite{zheng2017self}, trajectory basis (TB)\cite{park20153d}, high-pass filter (HPF)\cite{valmadre2012general} and the pseudo-triangulation approach in Sec. \ref{Sec:Optimization}. SEDL requires partial sequencing information. TB and HPF  require complete ground truth sequencing. 
%For fair comparison, w
We include a version of our method leveraging ground truth sequencing by enforcing  structural constraints on $\mathbb W$ similarly to HPF. %which we force only the two neighboring can be the adjacent frame when compute the $\mathbb{W}$.
%------------------------------------------------------------------------

\noindent {\bf Varying 2D noise}. We add white noise on the 2d observation with std. dev. from 1 to 5 pixels. The  parameters $\lambda_2$ and $\lambda_3$ are fixed as 0.0015 and 0.02.  Per Fig. \ref{fig:DiffNoise},  reconstruction accuracy degrades as the 2d observation error increases. Our method is  competitive with frameworks requiring sequencing info such as TB and HPF. 
%------------------------------------------------------------------------

\noindent {\bf  Varying frame rates}. We temporally downsample the motion capture datasets
%projected onto the virtual cameras 
and perform experiments at frames rates of 30 Hz, 15Hz and 7.5 Hz, without 2D observation noise. As shown in Fig. \ref{fig:DiffFreq}, without  sequencing info, our method outperforms  SEDL  for lower frame rates. Results for methods using full sequencing info are comparable.\\
%------------------------------------------------------------------------
\noindent {\bf Missing data}. We randomly decimate  10\% to 50\% of total 3D points before projection onto the virtual cameras. Reconstruction error comparisons are restricted to SEDL and TB, as other methods don't recover  missing joints. Per Fig. \ref{fig:DiffMissing}, our method has lower reconstruction error, across all missing data levels, compared to SEDL  with partial sequencing info and TB  with full sequencing info. \\
%------------------------------------------------------------------------
\noindent {\bf Non-uniform density}. We randomly drop 10\% to 50\% of total frames  from the motion  sequence. The reconstruction error increases disproportionately for the other methods compared to ours, as depicted in Fig. \ref{fig:DiffFrameDrop}. \\
%------------------------------------------------------------------------

 \begin{figure}
    \centering
    \fbox{\begin{subfigure}[ht!]{0.23\textwidth}
        \centering
        \includegraphics[scale = 0.08, trim={2.5cm 0.8cm 2cm 2cm},clip]{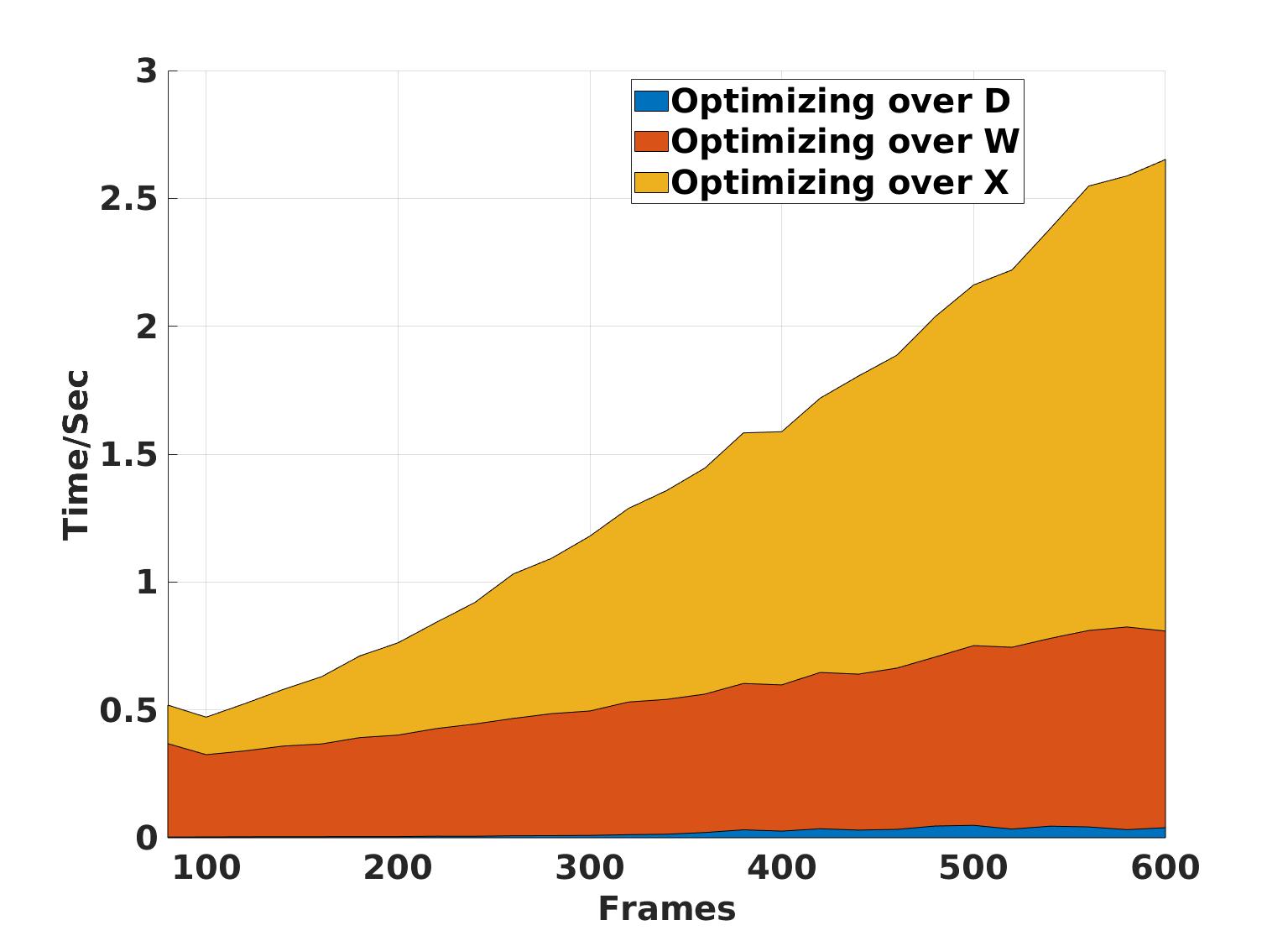}
        \caption{Single iteration run time \label{fig:Running time}}
    \end{subfigure}
    \begin{subfigure}[ht!]{0.22\textwidth}
        \centering
        \includegraphics[scale = 0.155, trim={4cm 0cm 4cm 0cm},clip]{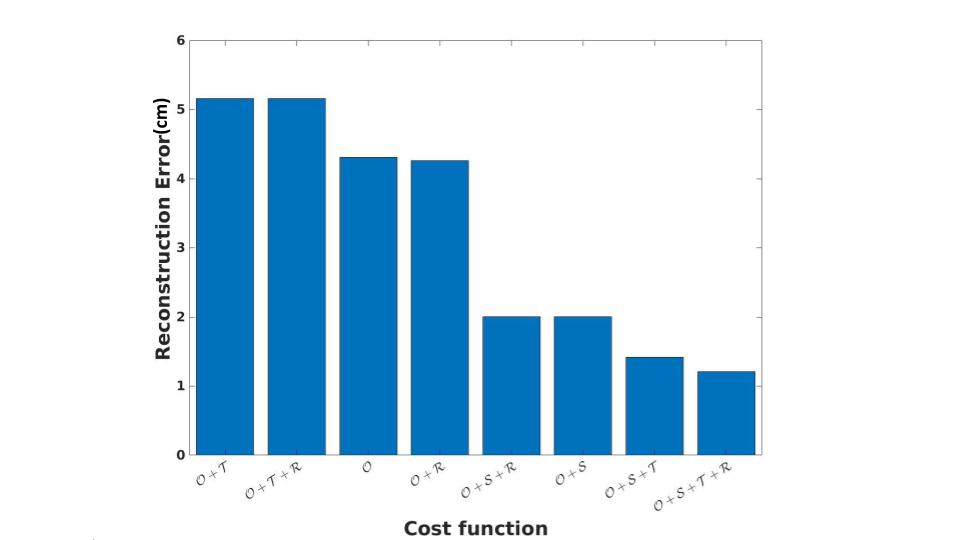}
        \caption{Ablation analysis}
        \label{fig:Ablation}
    \end{subfigure}}
    \caption{Optimization run time and cost function ablation}
\end{figure}

\noindent {\bf Execution run times}. \label{Runtime}
 Average run times for our Matlab implementation on an Intel i7-8700K CPU for optimizing each of our three variables are plotted in Fig. \ref{fig:Running time},   reconstructing $P=31$  features over a variable number of frames $N$. Time complexity for optimizing over $\mathbb{D}$ using an Active-Set method \cite{chen2014fast} is ${O}(min(3P,N)(PN+a^2))$, where $a$ is the number of non-zero values in the active-set. However, the number of estimation variables for this stage is only $N$. Optimizing $\mathbb{W}$ takes ${O}(min(3P,N)(PN+a^2)N)$ since we use the same solver for each row of $\mathbb{W}$.Optimizing over $\mathbb{X}$ is an unconstrained convex quadratic programming problem equating to solving a linear system of equations with time complexity of ${O}((NP)^3)$.
 Average running time  for minimizing either $\mathbb{X}$ or $\mathbb{W}$ are smaller due to the sparsity of  $\mathbb{W}$. 
 %We solved the quadratic problem using Alternate convex search. 
 Total number of iterations depends on initialization quality, reported experiments  ran an average of 62.26 iterations. \\
 %Our stopping criteria thresholds the sum of absolute difference values of $\mathbb{X}$  between successive iterations. 
\noindent {\bf Ablation Analysis. } \label{Ablation}
We analyze the contribution of the different terms in Eq. (\ref{LapOptOrigXL}) toward reconstruction accuracy for scenarios of moderate-to-high 2D noise levels. Fig. \ref{fig:Ablation} shows  results for multiple variants. The observation ray term $\mathcal O$ is common to all variants. Best performance is achieved by the instance optimizing over all geometric terms.

%------------------------------------------------------------------------
\subsection{Multi-view Video and Image Datasets} \label{Sec_RealData}
%------------------------------------------------------------------------
Experiments on imagery with known camera geometry include  Juggler\cite{ballan2010unstructured}, Climb \cite{park20103d} and Ski\cite{rhodin2018learning} datasets.  {\em We  unsynchronized images  by removing concurrent observations}, randomly selecting a single camera when multiple images shared a common timestamp. Timestamps were only used for  eliminating concurrency. 
For Juggler we use as 2D features the joint positions detected by  \cite{wei2016convolutional}.
For Climb and Ski we used the provided 2D feature tracks and  2D joint  detection locations, respectively. 
Fig. \ref{fig_MultiViewDatasets} illustrates our results  and describes the experimental setup.

\subsection{Application to Event Segmentation \label{Sec_EventSegmentation}}
We consider the case of dynamic reconstruction of spatially co-located, but temporally disjoint events captured in a single aggregated image set. For such scenario we obtain a Laplacian matrix describing  a graph with multiple connected components, one per each event. Importantly, for each component  we sequence its images and  reconstruct its dynamic 3D geometry. Spectral analysis of the Laplacian matrix visualizes the chain-like topology of each of these events/clusters, see Fig. \ref{fig_EventSegmentation} top right. 

\subsection{Application to Multi-Target Scenarios}
%Applying our framework in scenarios involving multiple targets interacting in a common environment is straightforward... as long as the instance identification problem is solved {\em a priori}. 
Given $M$ subjects observed in $N$ images, our aggregated shape representation $\mathbb X_{i,:} \in \mathbb R^{3MP}$  requires solving data associations of input 2D features among  $M$  subjects across $N$ images \cite {vo2018automatic}.  To this end, we leverage DLOE's event segmentation capabilities  (section \ref{Sec_EventSegmentation}) as follows: {\bf 1}) For each input $\mathcal{I}_n$, we create a proxy image $\tilde{\mathcal{I}}_q$ for each subject observed therein. 
%Hence,   2D features from each separate target are the only  features in a newly created image. 
{\bf 2}) Execute  DLOE  on the aggregated set of proxy images  $\{{\tilde{\mathcal{I}}_{q}| _{N \leq q \leq MN}}\}$ 
%where $N \leq q \leq MN$  
(each observing $P$ 3D points) to reconstruct each subject's motion as a distinct event. {\bf 3}) Associate 3D estimates of $\{{\tilde{\mathcal{I}}_q}\}$ based on their common ancestor $\mathcal{I}_n$, providing a coalesced spatio-temporal context for each reconstructed event.  {\bf 4}) Aggregate the 2D features of all sibling ${\tilde{\mathcal{I}}_q}$ into a single 2D shape representation, enforcing data associations from each event. {\bf 5}) Run  DLOE  on the aggregated representation over the $N$ original input images, to improve the decoupled event reconstructions from step 2.  %, using the previously estimated  3D geometry of each target-event as initialization. 
Fig. \ref{fig_MultiPerson} shows our workflow results for a two-target scenario.

%----------------------------------------------------------------------------

\section{Conclusion}
We presented a data-adaptive framework for the modeling of spatio-temporal relationships among visual data. Our tri-convex optimization framework  outperforms  state of the art methods for the challenging scenarios of decreasing and irregular temporal sampling. The generality of the formulation and internal data representations suggest robust dynamic 3D reconstruction as a data association framework for video.
%the geometric insights leveraged in our estimation are of general applicability.

{\small
\bibliographystyle{ieee_fullname}
\bibliography{egbib}
}

\end{document}